\documentclass{article}

\PassOptionsToPackage{numbers,compress}{natbib}

\usepackage[main, final]{neurips_2025}
\makeatletter
\renewcommand{\@noticestring}{}
\makeatother

\usepackage[utf8]{inputenc} 
\usepackage[T1]{fontenc}    
\usepackage{float}          
\usepackage{hyperref}       
\hypersetup{hidelinks}
\usepackage{booktabs}       
\usepackage{array}          
\usepackage{multirow}       
\usepackage{tabularx}       
\usepackage{etoolbox}       
\newcolumntype{Y}{>{\centering\arraybackslash}X}
\AtBeginEnvironment{table}{\small}
\AtBeginEnvironment{tabular}{\small}
\AtBeginEnvironment{tabularx}{\small}
\usepackage{graphicx}       
\usepackage{amsfonts}       
\usepackage{microtype}      
\usepackage{xcolor}         
\usepackage{colortbl}       
\definecolor{tablegroup}{RGB}{241,245,250}
\definecolor{tablegroupwarm}{RGB}{250,244,235}
\definecolor{tableours}{RGB}{230,239,249}
\usepackage{fontawesome5}   

\title{BWM: A Low-Cost High-Fidelity World Simulator for Robot Learning}

\author{\large BWM Team}

\makeatletter
\patchcmd{\@bottomtitlebar}{\vskip 0.09in}{\vskip 0.02in}{}{}
\patchcmd{\@maketitle}
  {\begin{tabular}[t]{c}\bf\rule{\z@}{24\p@}\@author\end{tabular}}
  {\begin{tabular}[t]{c}\bf\rule{\z@}{18\p@}\@author\end{tabular}}
  {}{}
\patchcmd{\@maketitle}
  {\vskip 0.3in \@minus 0.1in}
  {\vskip 0.03in \@minus 0.01in}
  {}{}
\makeatother

\begin{document}
\raggedbottom

\maketitle

\begin{center}
{\small
\faGithub\ Code: \href{https://github.com/boundless-large-model/boundless-world-model}{github.com/boundless-large-model/boundless-world-model}\\[-0.1em]
\href{https://huggingface.co/BLM-Lab/Boundless-World-Model}{\raisebox{-0.12em}{\includegraphics[height=1.05em]{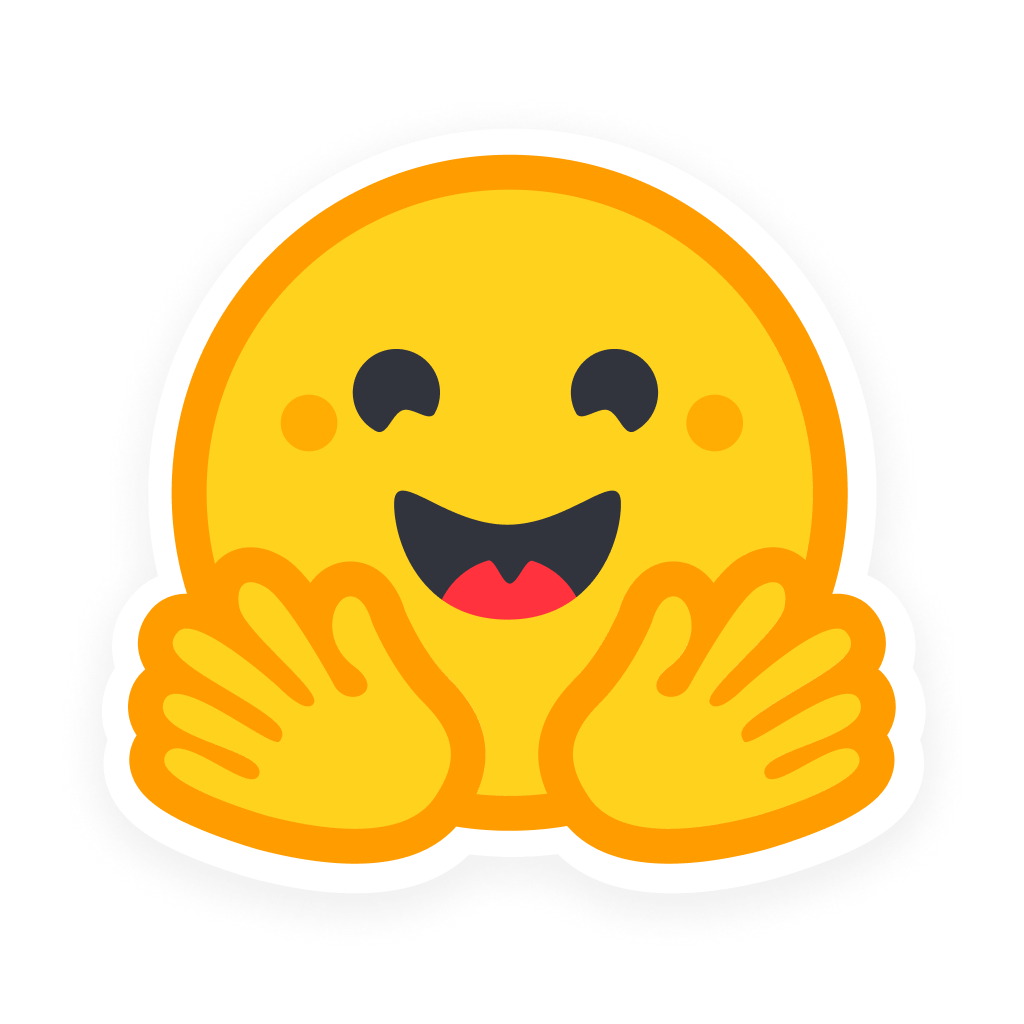}}}\ Model: \href{https://huggingface.co/BLM-Lab/Boundless-World-Model}{huggingface.co/BLM-Lab/Boundless-World-Model}
}
\end{center}
\vspace{-0.4em}

\begin{figure}[H]
    \centering
    \includegraphics[width=\linewidth]{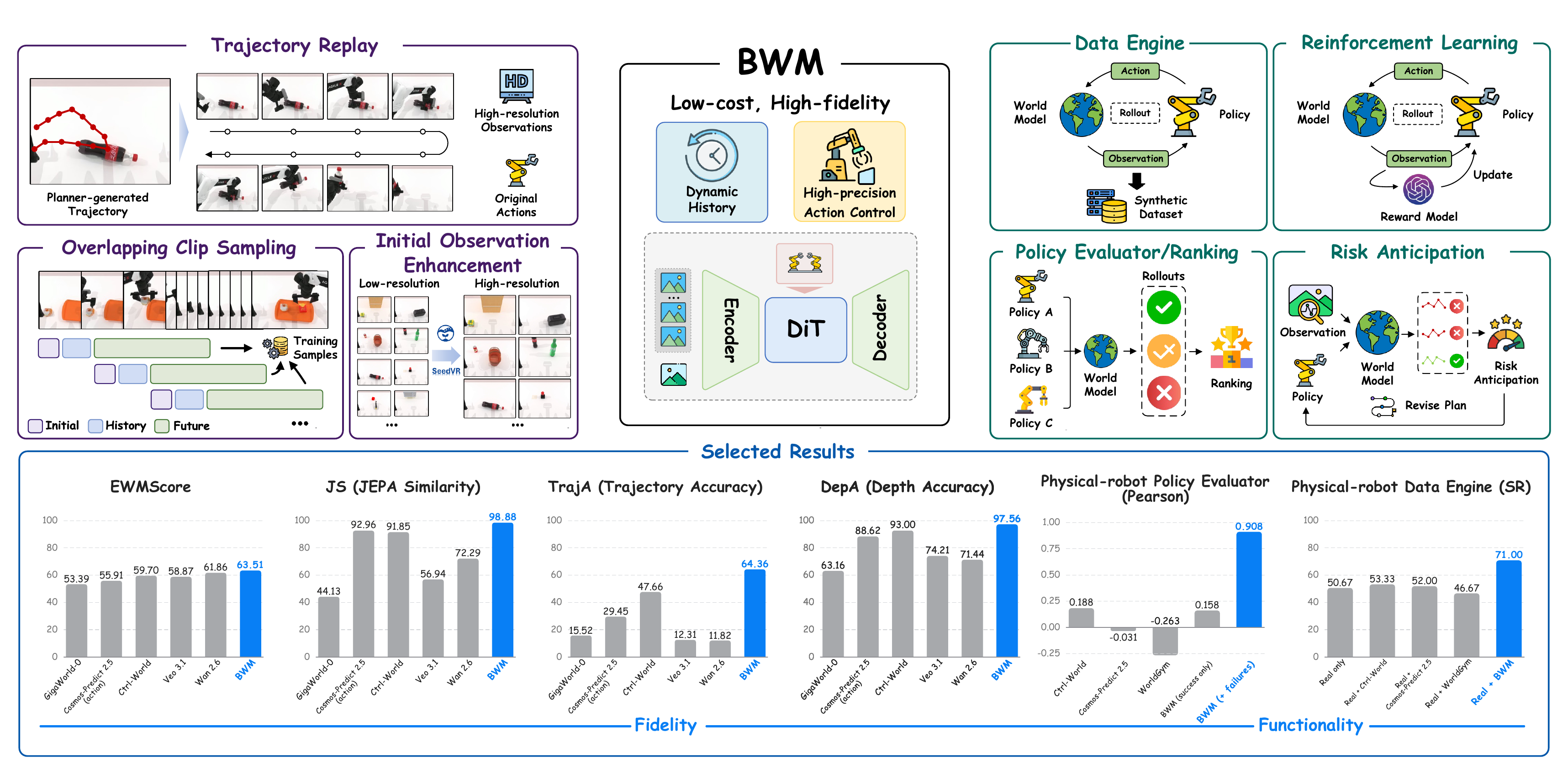}
    \caption{\textbf{Overview of BWM.} BWM combines action-aligned data construction, action-conditioned world simulation, and downstream robot-learning applications.}
    \label{fig:bwm_teaser}
\end{figure}

\begin{abstract}
Reliable robot learning requires a world simulator that can predict action consequences before execution on physical hardware, including risky and failure-prone outcomes.
Existing physics simulators require substantial asset construction and calibration and still face a sim-to-real gap, while video generators often lack precise control over their responses to fine-grained robot actions.
In this paper, we present the Boundless World Model (BWM), an open-source, low-cost, high-fidelity world simulator for robot manipulation.
BWM is an action-conditioned world model that combines initial-environment guidance, dynamic visual history, and temporally aligned robot-action conditioning for stateful autoregressive prediction of future observations.
We construct action-aligned training clips through trajectory replay, overlapping clip sampling, and initial-observation enhancement.
BWM serves as a data engine that augments imitation-learning data with action-aligned rollouts, and as a policy evaluator for closed-loop assessment, risk anticipation, and policy ranking.
Experiments on the WorldArena benchmark and physical robots demonstrate improved simulator fidelity and functional utility across the data-engine and policy-evaluator settings.
BWM ranks first overall in the WorldArena Challenge across Track~1 and its two Track~2 applications.
We release the BWM open-source ecosystem, including model checkpoints, training and inference code, and interfaces for data generation and policy evaluation.
\end{abstract}

\section{Introduction}

Recent VLA systems aim to generalize across tasks and embodiments, but achieving this goal still depends on heterogeneous supervision from real-robot trajectories, human videos, and synthetic robot data~\citep{geminirobotics2025,grootn12025}. Acquiring high-quality interaction data on hardware is costly, while collecting failures and corrective recoveries requires autonomous execution, human intervention, or explicit failure generation, making such data difficult to obtain safely at scale~\citep{recap2025,failsafe2025,recall2026}. Physics simulators can parallelize data generation and policy optimization. However, VLA policies learned or evaluated in these environments can degrade after transfer to physical robots because of discrepancies in appearance, sensing, dynamics, and control. Domain randomization improves robustness but does not eliminate this sim-to-real gap~\citep{simpler2024,robotwin22025,survey2026c}.

A learned robot world simulator provides a complementary source of interaction by predicting the visual consequences of candidate controls from recorded robot trajectories~\citep{survey2026a,survey2026b}. By learning appearance and dynamics directly from real-world interaction data, it can narrow the sim-to-real gap associated with physics simulators and reduce reliance on real-robot hardware. During policy learning, it can synthesize action-aligned trajectories for imitation learning and provide interactive rollouts for reinforcement learning. During deployment, it can evaluate a policy through closed-loop interaction and anticipate risky action consequences before physical execution. The simulator can generate video rollouts and evaluate policies without executing every interaction on physical hardware or fully reconstructing each setting in a physics engine. To serve these roles, the simulator must be responsive to actions and preserve task state, not merely generate plausible video.

Video world models~\citep{sora2024,moviegen2024,wan2025} have rapidly improved the generation of coherent scene evolution, and large-scale pretraining provides broad priors over appearance, motion, and visual dynamics. However, their generation interfaces do not expose the precise control variables required by a robot simulator. Interactive world models~\citep{gamengen2024,genie32025} introduce user control through keyboard events, navigation commands, camera motion, or high-level prompts. These low-dimensional inputs mainly govern viewpoint or global scene changes rather than the temporally aligned, high-precision actions needed to reproduce manipulation. Current robot action-conditioned world models~\citep{ctrlworld2025,worldgym2025} commonly adapt pretrained video generation backbones through post-training on large-scale robot datasets. Even with this adaptation, producing high-fidelity future rollouts with consistent responses to robot actions remains difficult.

We view a robot world simulator as a domain-aware adaptation of a general video world model, similar to configuring a specific simulation environment or a real-world experimental setup. General-purpose pretraining for video generation~\citep{sora2024,wan2025} provides reusable visual and physical priors for such simulators. Robot post-training adapts these priors to the scenes, embodiment, sensing setup, and action space of a target deployment. General-purpose robot post-training~\citep{cosmospredict2025,abotphysworld2026} seeks to support multiple embodiments and tasks within a shared model. However, variation in camera views, embodiments, and action spaces increases model capacity and training requirements, making it difficult for current systems to achieve both high-fidelity prediction and real-time simulation. We therefore adopt domain-specific robot post-training as a low-cost route to high-fidelity simulation, enabling rapid adaptation by retaining general video priors and focusing model capacity on the target robot and environment.

We introduce the \textbf{Boundless World Model (BWM)}, a low-cost, high-fidelity, action-conditioned world simulator. BWM introduces core improvements to both the data pipeline and model architecture. The pipeline builds action-aligned training data through trajectory replay, overlapping clip sampling, and initial-observation enhancement, while the model architecture combines initial-environment guidance, dynamic history-frame conditioning, and high-precision action control to autoregressively generate future observation chunks. We validate BWM as both a data engine and a policy evaluator in simulation and on physical robots. The data engine augments imitation-learning datasets with action-aligned rollouts, while the policy evaluator scores and ranks candidate policy models through closed-loop interaction. In the WorldArena Challenge~\citep{worldarena2026}, BWM ranks \textbf{second overall} and \textbf{first among open-source entries} in Track~1, \textbf{first} on the open-source Track~2 data-engine leaderboard, and \textbf{second} on the corresponding policy-evaluator leaderboard. On physical robots, policies trained with BWM-generated data reach \textbf{71.00\%} mean success, compared with \textbf{53.33\%} for the strongest evaluated world-simulator baseline, while the failure-inclusive evaluator achieves a Pearson correlation coefficient ($r$) of \textbf{0.908} with hardware outcomes. Our contributions are:

\begin{itemize}
    \item We introduce BWM, a low-cost, high-fidelity world simulator for robot learning. It supports synthetic data generation for imitation learning, interactive simulation for reinforcement learning, closed-loop policy evaluation, candidate policy ranking, and risk anticipation before physical execution.
    \item We jointly improve BWM's data pipeline and action-conditioned model architecture. The pipeline preserves action-observation alignment during data construction, while the model combines initial-environment guidance, dynamic visual history, and precise robot control for stateful autoregressive simulation.
    \item BWM ranks first overall when scores from WorldArena Track~1 and the two Track~2 applications are combined. On physical robots, BWM-generated data attains state-of-the-art downstream policy performance, while the closed-loop evaluator yields rankings that closely match hardware outcomes.
    \item We release the BWM open-source ecosystem, including model checkpoints, training and inference code, and interfaces for data generation and policy evaluation.
\end{itemize}

\section{Related Work}

\subsection{Video World Models}

Large-scale video models learn priors over appearance, motion, and scene dynamics from diverse video datasets. For example, Stable Video Diffusion~\citep{stablevideodiffusion2023}, HunyuanVideo~\citep{hunyuanvideo152025}, Seedance~\citep{seedance152025}, and Wan~\citep{wan2025} represent recent advances in large-scale video generation. Sora~\citep{sora2024} investigates scaled video generation as a route toward general-purpose simulation of physical and digital environments. The resulting models capture complex visual dynamics across a broad range of scenes.
Despite this progress, most video world models are optimized for open-ended generation and visual content synthesis. Their training objectives prioritize visually convincing future frames but do not explicitly model how control signals change the underlying world state. This lack of controllable transition modeling limits their direct use as interactive simulators. An interactive simulator should map external inputs to consistent state transitions and preserve temporal coherence across successive predictions.

\subsection{Action-Conditioned World Models}

\paragraph{Interactive Environments.}
Interactive world models generate visual environments that respond to user inputs. GameNGen~\citep{gamengen2024} predicts the next game frame from past frames and actions, enabling real-time interaction with a complex game environment. GameFactory~\citep{gamefactory2025} uses multi-phase training to decouple game-style learning from action control, preserving open-domain scene generalization while enabling action-controllable game video generation. Matrix-Game~\citep{matrixgame32026} combines camera-aware memory retrieval with multi-segment autoregressive distillation for real-time long-form generation. Genie~\citep{genie32025} generates dynamic worlds from text prompts that support real-time navigation and promptable world events.
These works focus on games or general navigable worlds. Their reported interfaces use player actions, pose and action inputs, navigation commands, or text events rather than robot action trajectories aligned with fine-grained robot--object interaction.

\paragraph{Robot Action Conditioning.}
Robot world models condition video prediction on temporally aligned, high-precision robot actions. IRASim~\citep{irasim2025} conditions video generation on historical observations and robot action trajectories, using a frame-level action-conditioning module within each transformer block to strengthen action--frame alignment. Cosmos-Predict~2.5~\citep{cosmospredict2025} adapts large-scale video pretraining to action-conditioned prediction through robot-specific post-training. HMA~\citep{maskedautoregression2025} uses heterogeneous pre-training from observations and action sequences across robot embodiments, domains, and tasks, then applies masked autoregression to generate quantized or soft tokens for video prediction. BridgeV2W~\citep{bridgev2w2026} converts coordinate-space actions into pixel-aligned embodiment masks rendered from URDF descriptions and camera parameters. ABot-PhysWorld~\citep{abotphysworld2026} converts robot commands into spatially structured action maps for action-conditioned video generation.
These systems commonly rely on large-scale video pretraining or heterogeneous robot pretraining, which increases the overall cost of building a world simulator.

\subsection{World Models for VLA}

\paragraph{Policy Optimization.}
World models support policy improvement by providing imagined interaction, reward, or value signals~\citep{survey2026a}. World4RL~\citep{world4rl2025} refines pretrained policies in imagined environments using sparse success signals. World-Env~\citep{worldenv2025} uses a VLM-guided instant reflector to provide continuous rewards and predict action termination. SRPO~\citep{srpo2025} uses successful rollouts from the current batch as self-references to assign progress-wise rewards to failed attempts. WoVR~\citep{wovr2026} reduces error accumulation through keyframe-initialized rollouts and maintains policy--simulator alignment through world model--policy co-evolution. Together, these methods provide a low-cost alternative to reinforcement learning through physical robot interaction.

\paragraph{Policy Evaluation.}
World models also provide a low-cost route for evaluating policies before hardware deployment. WorldGym~\citep{worldgym2025} estimates policy outcomes through action-conditioned rollouts and vision-language rewards. Scalable Policy Evaluation~\citep{scalableevaluation2025} studies whether learned video environments preserve policy rankings and reference performance estimates. Genie Envisioner~\citep{genieenvisioner2025} integrates learned simulation with policy evaluation and improvement. These systems make quantitative comparison possible without executing every candidate policy on hardware.

\paragraph{Data Engines.}
World-model data engines expand robot training data by generating new observations paired with corresponding actions. Ctrl-World~\citep{ctrlworld2025} combines multi-view prediction, pose-conditioned memory retrieval, and frame-level action conditioning, serving as a data engine that synthesizes successful trajectories for supervised policy improvement. DreamGen~\citep{dreamgen2025} synthesizes robot videos and recovers pseudo-actions through learned action models. GigaWorld-0~\citep{gigaworld2025} combines controllable video generation with 3D modeling and motion planning, while WM-DAgger~\citep{wmdagger2026} generates recovery data for iterative imitation learning. WristWorld~\citep{wristworld2025} augments missing wrist-view observations from other cameras. These methods provide scalable supplements to data collected on physical robots.

Existing studies often evaluate a single downstream task or functional role. Comprehensive evaluation remains limited, particularly protocols that jointly assess video quality across perceptual, physical, and action-response dimensions, measure multiple downstream functions, and cover both simulation and physical-robot settings.

\section{Data Pipeline}

The BWM training pipeline consists of trajectory replay, overlapping clip sampling, and initial-observation enhancement. Simulator replay is used to rerender training trajectories at a higher resolution. When replay is unavailable at inference time, only the initial observation is enhanced before rollout. These operations leave each action chunk and its temporal indexing unchanged.

\paragraph{Trajectory Replay.}

We use robot manipulation trajectories collected in the RoboTwin simulator~\citep{robotwin22025}. For each trajectory, we rerun the original planner-generated trajectory and render observations at the target resolution. Replay does not introduce new robot behavior. Its purpose is to collect higher-resolution observations along the original trajectory. Because the execution order and timestamps remain unchanged, the rerendered frames remain synchronized with the original absolute end-effector (EEF) actions.

\paragraph{Overlapping Clip Sampling.}

We extract temporally aligned training clips using overlapping sliding windows. Compared with non-overlapping segmentation, overlapping windows preserve transitions that would otherwise be split by fixed clip boundaries and expose them under different temporal contexts. This sampling strategy also better matches the sliding context window used during autoregressive inference. Each clip contains an initial-environment observation, a dynamic history window, a future observation window, and the temporally aligned EEF action chunk. During inference, the initial-environment observation remains fixed, while the history window and action chunk advance after each rollout step.

\paragraph{Initial-Observation Enhancement.}

When simulator replay is unavailable at inference time, we restore the initial-environment observation using SeedVR-2~\citep{seedvr22026} before rollout.

\section{Boundless World Model}

\begin{figure}[t]
    \centering
    \includegraphics[width=\linewidth]{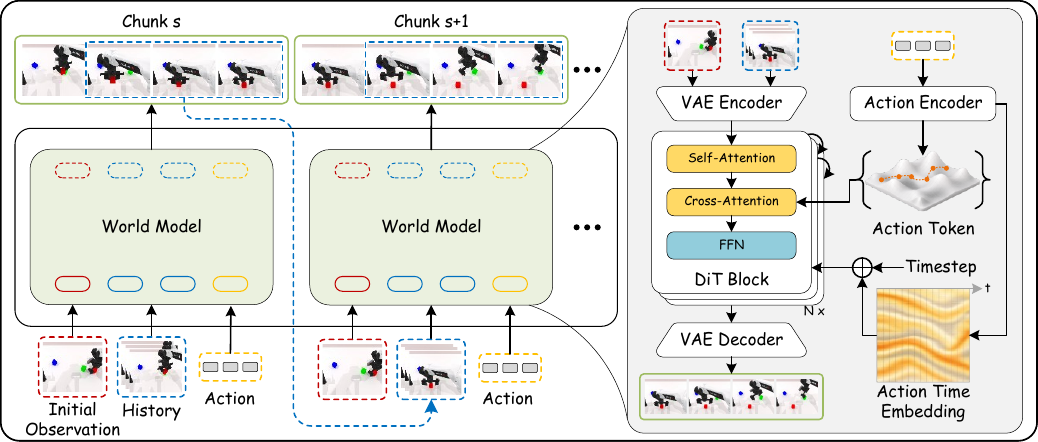}
    \caption{\textbf{Overview of BWM.} BWM autoregressively predicts future observation chunks from the initial observation, dynamic history, and action chunk. Robot actions enter the video diffusion backbone as cross-attention tokens and action-conditioned timestep embeddings.}
    \label{fig:bwm_framework}
\end{figure}

\subsection{Problem Setup}

Figure~\ref{fig:bwm_framework} presents the overall architecture of BWM. BWM formulates world simulation as action-conditioned prediction in observation space. Its predictive core estimates how the visual state of a manipulation scene changes under a candidate sequence of robot controls. Let $x_i\in\mathcal{X}$ be the robot-view observation at step $i$. In the reported implementation, $a_i\in\mathbb{R}^{d_a}$ denotes the absolute end-effector (EEF) pose command recorded at the same step, where $d_a$ is the action dimension. At rollout step $t$, the dynamic history is $\mathbf{h}_t=(x_{t-H+1},\ldots,x_t)$, where $H$ is the history-window length and the entries may be observed or previously generated. Given the initial-environment observation $x_0$, the dynamic history $\mathbf{h}_t$, and the candidate action chunk $\mathbf{a}_{t+1:t+K}$, BWM parameterizes the distribution of the next $K$ observations:
\begin{equation}
    p_{\theta}(\mathbf{x}_{t+1:t+K} \mid x_0, \mathbf{h}_t, \mathbf{a}_{t+1:t+K}).
    \label{eq:bwm_prediction}
\end{equation}
The initial environment and dynamic history define the visual state, while the action chunk specifies the transition to be simulated. Repeatedly feeding predicted observations back into $\mathbf{h}_t$ turns the chunk predictor into a learned visual simulator. Effective rollouts should produce visual changes that respond to the supplied actions while preserving scene consistency across autoregressive steps.

\subsection{Action-Conditioned Rollout Model}

BWM adapts a pretrained video diffusion model through a task-specific action interface. For each action dimension, we use the 1st and 99th percentiles of the training data, denoted by $p_1$ and $p_{99}$, as the normalization bounds and map the clipped values to $[\ell_a,u_a]$. The action interface injects the same action chunk through two complementary paths. The frame-level path encodes each action independently and supplies the resulting tokens to cross-attention, preserving fine temporal control. The latent-level path temporally aggregates and projects the action sequence to align with the compressed video latents. The resulting representations are added to the timestep embeddings to modulate AdaLN. Cross-attention provides framewise action cues, while AdaLN conditions denoising at the latent temporal resolution.

Each training clip contains an initial-environment observation, $H$ history observations, $K$ future targets, and an action sequence under the same temporal indexing. Let $z^0$ and $z^{\mathrm{hist}}$ encode the initial environment and history, and let $z^{\mathrm{fut}}$ encode the future target. Following standard flow matching, we sample a scheduler timestep $\tau$ and construct $z^{\mathrm{fut}}_\tau$ with its velocity target $v^{\mathrm{fut}}_\tau$~\citep{flowmatching2023}. The initial-environment latent $z^0$ remains clean, while the history latents receive a low-level perturbation, $\widetilde z^{\mathrm{hist}}=(1-\sigma_h)z^{\mathrm{hist}}+\sigma_h\epsilon^{\mathrm{hist}}$, where $\epsilon^{\mathrm{hist}}\sim\mathcal{N}(0,I)$ and $\sigma_h$ is the history-noise scale. The resulting model input is $\widetilde z_\tau=[z^0,\widetilde z^{\mathrm{hist}},z^{\mathrm{fut}}_\tau]$. The future-only flow-matching objective is
\begin{equation}
    \mathcal{L}_{\mathrm{future}}
    = \mathbb{E}\!\left[
      w(\tau)\left\|v^{\mathrm{fut}}_{\theta}
      (\widetilde z_\tau,\mathbf{a},\tau)-v^{\mathrm{fut}}_\tau
    \right\|_2^2\right],
    \label{eq:future_loss}
\end{equation}
where $w(\tau)$ is the scheduler-defined timestep weight, and the expectation is taken over the training data, sampled noise, and scheduler timesteps.

Autoregressive inference uses the same history perturbation. At every denoising step, BWM replaces the conditional prefix with the clean initial-environment latent and fixed perturbed history. After decoding, the model discards the conditional frames, appends only $\mathbf{x}_{t+1:t+K}$ to the rollout, and updates $\mathbf{h}_t$ before processing the next action chunk. Repeating this process produces a stateful visual rollout while the initial-environment guidance preserves scene identity.

\subsection{Simulator Interfaces}

The learned visual simulator supports two interfaces through the same action-conditioned rollout model. The data engine generates action-aligned trajectories for imitation learning. The policy evaluator performs closed-loop assessment, anticipates risky outcomes, and ranks candidate policy models before hardware deployment.

\paragraph{Data Engine.}
The model receives aligned observation and action context, then generates future observation chunks offline. Each generated chunk remains paired with the action sequence that produced it, allowing the resulting trajectories to augment imitation-learning data.

\paragraph{Policy Evaluator.}
The same closed-loop protocol is used with a fixed policy and without policy updates. A task-specific success criterion converts each generated trajectory into an outcome score. Aggregated scores rank candidate policy models and can be compared with reference-simulator or hardware performance. During online evaluation, successful and failed rollouts reveal both expected performance and the risk associated with proposed action sequences before hardware execution.

\section{Experiments}

\subsection{Experimental Setup}

\paragraph{Evaluation Scope.}
We organize the evaluation around five research questions.
\begingroup
\setlength{\leftmargini}{1.6em}
\begin{itemize}
    \item \textbf{RQ1. Simulator Fidelity.} How well does BWM preserve visual appearance, interaction physics, spatial structure, and action-conditioned behavior?
    \item \textbf{RQ2. Data Engine Utility.} Can BWM-generated trajectories improve downstream policy performance?
    \item \textbf{RQ3. Policy Evaluation Fidelity.} Can closed-loop rollouts reproduce policy performance measured in a reference physics simulator?
    \item \textbf{RQ4. Physical-Robot Functionality.} Do the data engine and policy evaluator remain effective on physical robots?
    \item \textbf{RQ5. Design Analysis.} How do context design, action conditioning, and data processing affect simulator fidelity?
\end{itemize}
\endgroup
We answer these questions using the WorldArena benchmark~\citep{worldarena2026} and a physical-robot evaluation across six manipulation tasks on the ARX X5 platform.

\paragraph{Fidelity Evaluation.}
WorldArena defines 16 generation metrics grouped into visual quality, motion quality, content consistency, physics adherence, 3D accuracy, and controllability.
All metrics are normalized to a 0 to 100 scale, and EWMScore is their arithmetic mean.
The benchmark uses 2,000 RoboTwin~2.0 videos for training and 500 for testing across 50 task scenarios~\citep{worldarena2026}.
The groups distinguish perceptual quality from scene consistency, physical interaction, spatial structure, and sensitivity to the supplied action condition.
We apply this evaluation framework to both simulated and physical-robot rollouts.

\paragraph{Functional Evaluation.}
The data-engine studies measure policy success after using world-model trajectories for policy training.
The simulation protocol compares alternative training sources, while the physical protocol adds model-generated trajectories to real data.
For policy evaluation, a fixed policy predicts an action chunk from the current observation, the world model predicts the next observation chunk under that action, and the generated observation is returned to the policy.
Repeating these steps forms a closed-loop rollout.
In simulation, we evaluate policy ranking by computing the Pearson correlation between performance estimated from rollouts and performance measured in the physics simulator across policy variants, after aggregating each variant's performance over all tasks.
On the physical robot, we evaluate task-level consistency for a single policy using success-rate errors and the Pearson correlation between rollout estimates and hardware outcomes across tasks.

\paragraph{Baselines.}
For the simulation study, baseline performance values are taken from WorldArena~\citep{worldarena2026}.
General video generators include Wan~2.2~\citep{wan2025}, CogVideoX~\citep{cogvideox2024}, Veo~3.1~\citep{worldarena2026}, and Wan~2.6~\citep{wan2025}.
Text-conditioned world models include GigaWorld-0~\citep{gigaworld2025}, TesserAct~\citep{tesseract2025}, RoboMaster~\citep{worldarena2026}, Vidar~\citep{vidar2025}, Cosmos-Predict~2.5 (text)~\citep{cosmospredict2025}, and WoW~\citep{wow2025}.
Action-conditioned world models include Genie Envisioner~\citep{genieenvisioner2025}, Cosmos-Predict~2.5 (action)~\citep{cosmospredict2025}, IRASim~\citep{irasim2025}, and Ctrl-World~\citep{ctrlworld2025}.
We select this set by considering model availability and reproducibility.
The simulation data-engine and policy-evaluator experiments use the protocols and baseline results reported by WorldArena~\citep{worldarena2026}.
For the physical-robot study, we reproduce Cosmos-Predict~2.5~\citep{cosmospredict2025}, Ctrl-World~\citep{ctrlworld2025}, and WorldGym~\citep{worldgym2025} from their released implementations and fine-tune them on data from our experimental scenes.

\paragraph{Implementation Details.}
BWM is initialized from the Wan2.2-TI2V-5B video diffusion model~\citep{wan2025}, implemented in PyTorch, and trained on four compute nodes, each equipped with eight NVIDIA A800 GPUs. We use a per-GPU batch size of 1 and accumulate gradients over four steps, giving an effective batch size of 128. The WorldArena submission model is trained for 12,000 optimization steps, requiring approximately 929 GPU-hours. The physical-robot model covers fewer tasks and therefore uses a shorter training schedule. The reported configuration uses $d_a=14$, $H=8$, and $K=72$. The action interface prepends $P=3$ boundary actions and groups actions with $G=4$, aligning the resulting tokens with the temporally compressed video latents.

\paragraph{Evaluation Details.}
\begingroup
\setlength{\leftmargini}{1.6em}
\begin{itemize}
    \item \textbf{Simulation fidelity.} We evaluate 500 local validation sequences, with 10 sequences for each of 50 tasks. Each sequence provides one initial frame and the complete subsequent action sequence.
    \item \textbf{Simulation data engine.} We evaluate \textit{adjust bottle} and \textit{click bell}. For each task, every world model generates 25 synthetic trajectories for training a $\pi_{0.5}$ policy, giving 50 trajectories across the two tasks.
    \item \textbf{Simulation policy evaluator.} We use five $\pi_{0.5}$ policies provided by WorldArena, corresponding to 10\%, 20\%, 30\%, 50\%, and 100\% of the policy-training data. Each policy is rolled out from 10 initial frames per task across 50 tasks.
    \item \textbf{Physical-robot fidelity.} We evaluate 10 validation sequences per task across six tasks, giving 60 sequences in total.
    \item \textbf{Physical-robot data engine.} For each of six tasks, the reference MOTIF policy~\citep{motif2026} is trained on 50 real trajectories. Each augmented MOTIF policy adds 40 trajectories generated by one world model, and its success rate is measured over 25 hardware trials.
    \item \textbf{Physical-robot policy evaluator.} We evaluate a fixed MOTIF policy~\citep{motif2026} through 25 closed-loop rollouts for each of six tasks and compare the estimated success rates with hardware outcomes using success-rate errors and Pearson correlation.
\end{itemize}
\endgroup

\begin{figure}[H]
    \centering
    \includegraphics[width=\linewidth]{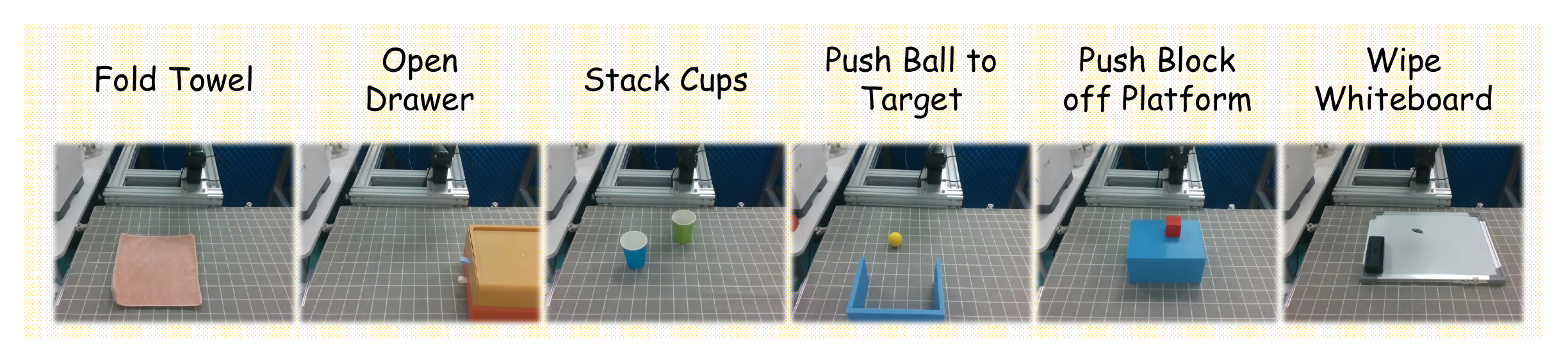}
    \caption{Physical-robot task scenarios.}
    \label{fig:physical_robot_tasks}
\end{figure}

\paragraph{Experimental Scenarios.}
The simulation setting covers 50 tasks from the open-source RoboTwin~2.0 suite used by WorldArena~\citep{robotwin22025,worldarena2026}.
The physical-robot setting covers \textit{Fold Towel}, \textit{Open Drawer}, \textit{Stack Cups}, \textit{Push Ball to Target}, \textit{Push Block off Platform}, and \textit{Wipe Whiteboard} on the ARX X5 platform, as shown in Figure~\ref{fig:physical_robot_tasks}.
\begingroup
\setlength{\leftmargini}{1.6em}
\begin{itemize}
    \item \textbf{Fold Towel.} The robot grasps the upper-right corner of a pink towel, folds it diagonally, and aligns the grasped corner with the lower-left corner.
    \item \textbf{Open Drawer.} The robot grasps the handle of the uppermost drawer and pulls the drawer horizontally outward.
    \item \textbf{Stack Cups.} The robot picks up a blue paper cup and stacks it on a green cup.
    \item \textbf{Push Ball to Target.} The robot pushes a yellow ball with its end effector until the ball rolls into a blue box.
    \item \textbf{Push Block off Platform.} The robot pushes a red block pre-positioned on the tabletop until it falls over the edge.
    \item \textbf{Wipe Whiteboard.} The robot grasps an eraser and moves the eraser back and forth across the whiteboard to remove the writing.
\end{itemize}
\endgroup

\subsection{Simulation Performance}

\subsubsection{Simulator Fidelity}

WorldArena compares models with different availability and conditioning interfaces.
Veo~3.1 and Wan~2.6 represent closed-source commercial video generators, while Wan~2.2 and CogVideoX provide open-source general video baselines.
The robot-oriented baselines include text-conditioned models such as GigaWorld-0 and WoW, together with action-conditioned world simulators, including Genie Envisioner, Cosmos-Predict~2.5 (action), IRASim, and Ctrl-World.
Tables~\ref{tab:sim_video_quality_worldarena} and~\ref{tab:sim_video_quality_physical_control} jointly report general generation quality and robot-centric measures of interaction, geometry, and controllability.

\begingroup
\let\originaltable\table
\let\endoriginaltable\endtable
\renewenvironment{table}[1][]{\originaltable[H]}{\endoriginaltable}
\begin{table}[!t]
\centering
\caption{Simulation fidelity on the WorldArena benchmark~\citep{worldarena2026}, covering EWMScore, visual quality, motion quality, and content consistency. EWMScore averages all 16 normalized submetrics. Abbreviations are EWM: EWMScore, IQ: Image Quality, AQ: Aesthetic Quality, JS: JEPA Similarity, DD: Dynamic Degree, FS: Flow Score, MS: Motion Smoothness, SC: Subject Consistency, BC: Background Consistency, and PC: Photometric Consistency. Best and second-best values are shown in \textbf{bold} and \underline{underlined}, respectively.}
\label{tab:sim_video_quality_worldarena}
{\renewcommand{\arraystretch}{1.10}
\setlength{\tabcolsep}{1pt}
\begin{minipage}{0.98\linewidth}
\begin{tabularx}{\linewidth}{>{\raggedright\arraybackslash}p{0.28\linewidth}*{10}{Y}}
\toprule
\multicolumn{1}{c}{\multirow{2}{*}{Model}} & \multicolumn{1}{c}{\makebox[0pt][c]{Overall}} & \multicolumn{3}{c}{Visual Quality} & \multicolumn{3}{c}{Motion Quality} & \multicolumn{3}{c}{Content Consistency} \\
\cmidrule(lr){2-2}\cmidrule(lr){3-5}\cmidrule(lr){6-8}\cmidrule(lr){9-11}
& EWM & IQ & AQ & JS & DD & FS & MS & SC & BC & PC \\
\midrule
GigaWorld-0 & 53.39 & 50.41 & 39.91 & 44.13 & 67.09 & 31.18 & 78.11 & 73.03 & 85.63 & 17.56 \\
Genie Envisioner & 43.65 & 23.05 & 32.89 & 33.40 & \underline{69.30} & 8.55 & 69.66 & 77.60 & 90.24 & 20.06 \\
TesserAct & 53.23 & 33.22 & \underline{45.90} & 45.79 & 51.50 & 24.47 & 75.79 & 82.50 & \textbf{92.38} & 24.91 \\
RoboMaster & 51.84 & 34.87 & 38.42 & 29.66 & 61.24 & 14.84 & 69.40 & 82.95 & \underline{91.23} & 33.56 \\
Vidar & 51.60 & 41.45 & 40.68 & 56.08 & 27.67 & 14.26 & \underline{79.73} & 76.29 & 83.00 & 23.50 \\
Cosmos-Predict 2.5 (text) & 50.81 & \underline{66.68} & 45.01 & 31.26 & 59.11 & \underline{43.02} & 78.82 & 74.88 & 85.11 & 13.83 \\
Cosmos-Predict 2.5 (action) & 55.91 & 44.89 & 35.76 & 92.96 & 39.94 & 5.73 & 71.00 & 81.97 & 88.94 & 35.28 \\
WoW & 54.88 & 45.87 & 38.68 & 74.40 & 46.08 & 27.06 & 76.92 & 81.61 & 90.25 & 21.70 \\
Ctrl-World & 59.70 & 35.22 & 38.93 & 91.85 & 42.57 & 34.49 & 73.77 & \textbf{84.11} & 90.57 & 17.29 \\
Wan 2.2 & 54.54 & 38.84 & 39.63 & 75.75 & 43.49 & 12.69 & 70.19 & \underline{83.88} & 90.42 & \textbf{47.76} \\
CogVideoX & 57.90 & 35.82 & 37.77 & \underline{93.84} & 31.66 & 21.89 & 73.91 & 80.83 & 87.73 & \underline{35.80} \\
IRASim & 58.12 & 34.89 & 36.23 & 93.30 & 41.39 & 20.83 & 70.52 & 83.12 & 90.68 & 35.22 \\
Veo 3.1 & 58.87 & 66.05 & \textbf{46.32} & 56.94 & 54.50 & 13.96 & 69.89 & 78.78 & 87.10 & 32.47 \\
Wan 2.6 & \underline{61.86} & \textbf{68.24} & 44.33 & 72.29 & \textbf{74.21} & \textbf{45.32} & \textbf{85.39} & 75.17 & 86.87 & 19.04 \\
\rowcolor{tableours}[\tabcolsep][\tabcolsep]\textbf{BWM} & \textbf{63.51} & 48.42 & 40.07 & \textbf{98.88} & 31.61 & 19.99 & 73.45 & 81.44 & 88.28 & 28.71 \\
\bottomrule
\end{tabularx}
\end{minipage}
}
\end{table}

\begin{table}[!t]
\centering
\caption{Simulation fidelity on the WorldArena benchmark~\citep{worldarena2026}, covering physics adherence, 3D accuracy, and controllability. Abbreviations are IntQ: Interaction Quality, TrajA: Trajectory Accuracy, DepA: Depth Accuracy, Persp: Perspectivity, InstF: Instruction Following, SemA: Semantic Alignment, and ActF: Action Following.}
\label{tab:sim_video_quality_physical_control}
{\renewcommand{\arraystretch}{1.10}
\begin{minipage}{0.98\linewidth}
\begin{tabularx}{\linewidth}{>{\raggedright\arraybackslash}p{0.28\linewidth}*{7}{Y}}
\toprule
\multicolumn{1}{c}{\multirow{2}{*}{Model}} & \multicolumn{2}{c}{\makebox[0pt][c]{Physics Adherence}} & \multicolumn{2}{c}{3D Accuracy} & \multicolumn{3}{c}{Controllability} \\
\cmidrule(lr){2-3}\cmidrule(lr){4-5}\cmidrule(lr){6-8}
& IntQ & TrajA & DepA & Persp & InstF & SemA & ActF \\
\midrule
GigaWorld-0 & 53.68 & 15.52 & 63.16 & 75.96 & 61.56 & 85.91 & \underline{11.34} \\
Genie Envisioner & 20.52 & 6.79 & 86.63 & 52.84 & 20.28 & 85.44 & 1.09 \\
TesserAct & 58.00 & 13.96 & 71.59 & 79.20 & 61.52 & 87.83 & 3.11 \\
RoboMaster & 53.64 & 11.58 & 83.35 & 75.88 & 57.72 & 87.61 & 3.52 \\
Vidar & 53.48 & 19.28 & 78.72 & 75.92 & 59.12 & 88.26 & 8.19 \\
Cosmos-Predict 2.5 (text) & 38.72 & 8.16 & 70.51 & 79.64 & 26.64 & 77.33 & \textbf{14.18} \\
Cosmos-Predict 2.5 (action) & 55.00 & 29.45 & 88.62 & 76.44 & 58.40 & 88.79 & 1.33 \\
WoW & 55.64 & 20.58 & 72.83 & 76.72 & 56.92 & 88.42 & 4.34 \\
Ctrl-World & 62.12 & \underline{47.66} & 93.00 & 79.60 & 72.72 & 89.12 & 2.10 \\
Wan 2.2 & 51.84 & 16.27 & 77.68 & 76.60 & 53.76 & 88.77 & 5.12 \\
CogVideoX & 59.40 & 35.26 & 90.97 & 78.28 & 72.68 & \underline{89.77} & 0.76 \\
IRASim & 56.56 & 36.39 & \underline{93.12} & 77.88 & 66.04 & 88.49 & 5.26 \\
Veo 3.1 & \textbf{78.72} & 12.31 & 74.21 & \underline{82.76} & \textbf{93.28} & 86.07 & 8.52 \\
Wan 2.6 & 72.80 & 11.82 & 71.44 & 80.32 & \underline{85.36} & 87.28 & 9.92 \\
\rowcolor{tableours}[\tabcolsep][\tabcolsep]\textbf{BWM} & \underline{73.76} & \textbf{64.36} & \textbf{97.56} & \textbf{93.08} & 82.48 & \textbf{90.59} & 3.42 \\
\bottomrule
\end{tabularx}
\end{minipage}
}
\end{table}

\endgroup

BWM ranks first overall with the \textbf{highest EWMScore of 63.51}.
It exceeds the highest-scoring closed-source video generator (Wan~2.6) by 1.65 and the highest-scoring open-source general video generator (CogVideoX) by 5.61.
Among action-conditioned world simulators, BWM improves over the strongest baseline (Ctrl-World) by 3.81 in EWMScore and 16.70 in trajectory accuracy.
BWM also records the best JEPA similarity, trajectory accuracy, depth accuracy, perspectivity, and semantic alignment, demonstrating clear advantages on key simulator metrics for scene consistency, robot-trajectory prediction, and 3D spatial fidelity.

\subsubsection{Data Engine}

The WorldArena benchmark tests whether action-aligned generated trajectories improve a downstream $\pi_{0.5}$ policy on \textit{adjust bottle} and \textit{click bell}.
Table~\ref{tab:sim_data_engine} reports task success rates under this data-engine setting.
Each world model supplies 25 synthetic trajectories for each of the two tasks for $\pi_{0.5}$ training, and the resulting policy is evaluated over 100 executions of each task.

\begingroup
\let\originaltable\table
\let\endoriginaltable\endtable
\renewenvironment{table}[1][]{\originaltable[H]}{\endoriginaltable}
\begin{table}[!t]
\centering
\caption{Data-engine performance on the WorldArena benchmark~\citep{worldarena2026}, covering downstream policy success on \textit{adjust bottle} and \textit{click bell}. Each reported method supplies 25 synthetic trajectories per task for $\pi_{0.5}$ training.}
\label{tab:sim_data_engine}
{\renewcommand{\arraystretch}{1.10}
\begin{tabular}{lccc}
\toprule
Model & \textit{adjust bottle} & \textit{click bell} & Avg. \\
\midrule
$\pi_{0.5}$ policy model (zero-shot) & 2.00 & 5.00 & 3.50 \\
$\pi_{0.5}$ policy model (trained with real data) & \underline{77.00} & 66.00 & \underline{71.50} \\
Genie Envisioner & 7.00 & 21.00 & 14.00 \\
TesserAct & 1.00 & 35.00 & 18.00 \\
RoboMaster & 7.00 & 68.00 & 37.50 \\
Vidar & 13.00 & 53.00 & 33.00 \\
WoW & 45.00 & \underline{71.00} & 58.00 \\
Wan 2.2 & 15.00 & 41.00 & 28.00 \\
\rowcolor{tableours}[\tabcolsep][\tabcolsep]\textbf{BWM} & \textbf{98.00} & \textbf{91.00} & \textbf{94.50} \\
\bottomrule
\end{tabular}
}
\end{table}

\endgroup

BWM achieves the \textbf{highest average success rate of 94.50\%}.
This result is 23.00 above the policy trained with real data and 36.50 above WoW, the strongest alternative world-simulator data source.
BWM reaches 98.00\% on \textit{adjust bottle} and 91.00\% on \textit{click bell}, giving the best result on both tasks.
The two-task comparison shows that BWM-generated trajectories provide effective supervision for downstream policy training.

\subsubsection{Policy Evaluator}

For closed-loop evaluation, we ask whether simulator rollouts preserve the policy-performance ordering measured by the RoboTwin simulator.
WorldArena provides five $\pi_{0.5}$ policies with different capability levels for evaluation in each action-conditioned world model and uses a vision-language model to judge rollout success.
Table~\ref{tab:sim_policy_evaluator} reports the Pearson correlation between the resulting success rates and RoboTwin outcomes across these policies.

\begingroup
\let\originaltable\table
\let\endoriginaltable\endtable
\renewenvironment{table}[1][]{\originaltable[H]}{\endoriginaltable}
\begin{table}[!t]
\centering
\caption{Policy-evaluator performance on the WorldArena benchmark~\citep{worldarena2026}, covering Pearson correlation with RoboTwin simulator outcomes across policies with different capability levels.}
\label{tab:sim_policy_evaluator}
{\renewcommand{\arraystretch}{1.10}
\begin{tabular}{lc}
\toprule
World Simulator & Pearson $r$ $\uparrow$ \\
\midrule
Cosmos-Predict 2.5 (action) & 0.483 \\
IRASim & 0.658 \\
RoboScape & 0.863 \\
WoW & 0.959 \\
Ctrl-World & \textbf{0.986} \\
\rowcolor{tableours}[\tabcolsep][\tabcolsep]\textbf{BWM} & \underline{0.978} \\
\bottomrule
\end{tabular}
}
\end{table}

\endgroup

BWM reaches a {\bfseries\boldmath Pearson correlation of $r=0.978$}, within 0.008 of Ctrl-World and 0.019 above WoW.
It is one of only two evaluated world simulators with a correlation above 0.97.
This high agreement shows that BWM supports reliable policy ranking through closed-loop rollouts.

\subsection{Physical-Robot Performance}

\subsubsection{Real-Video Fidelity}

On physical-robot videos, we examine whether BWM retains fidelity across third-person and wrist-camera views.
Tables~\ref{tab:real_video_quality} and~\ref{tab:real_video_quality_physical_control} compare Cosmos-Predict~2.5, Ctrl-World, WorldGym, and BWM using the available perceptual, physical, spatial, and controllability metrics.
Action following is unavailable in both views, and wrist-view trajectory accuracy is also unavailable.

\begingroup
\let\originaltable\table
\let\endoriginaltable\endtable
\renewenvironment{table}[1][]{\originaltable[H]}{\endoriginaltable}
\begin{table}[!t]
\centering
\caption{Simulator performance on physical-robot videos, covering EWMScore, visual quality, motion quality, and content consistency from third-person and wrist-camera views. EWMScore averages 15 metrics for the third-person view and 14 for the wrist view.}
\label{tab:real_video_quality}
{\renewcommand{\arraystretch}{1.10}
\setlength{\tabcolsep}{1pt}
\begin{minipage}{0.98\linewidth}
\begin{tabularx}{\linewidth}{>{\raggedright\arraybackslash}p{0.23\linewidth}*{10}{Y}}
\toprule
\multicolumn{1}{c}{\multirow{2}{*}{Model}} & \multicolumn{1}{c}{\makebox[0pt][c]{Overall}} & \multicolumn{3}{c}{Visual Quality} & \multicolumn{3}{c}{Motion Quality} & \multicolumn{3}{c}{Content Consistency} \\
\cmidrule(lr){2-2}\cmidrule(lr){3-5}\cmidrule(lr){6-8}\cmidrule(lr){9-11}
& EWM & IQ & AQ & JS & DD & FS & MS & SC & BC & PC \\
\midrule
\rowcolor{tablegroupwarm}[\tabcolsep][\tabcolsep]\multicolumn{11}{l}{\emph{Third-person view}} \\
Cosmos-Predict 2.5 & \underline{66.53} & \textbf{58.62} & \underline{37.32} & \underline{73.01} & \textbf{17.27} & \underline{4.10} & \underline{50.50} & \textbf{71.01} & \textbf{71.09} & \textbf{80.29} \\
Ctrl-World & 46.44 & 56.44 & 35.06 & 18.71 & 16.28 & \textbf{7.62} & \textbf{51.74} & 61.51 & 62.44 & 26.11 \\
WorldGym & 59.06 & 48.30 & 37.06 & 56.26 & 13.01 & 3.14 & 50.31 & 52.79 & 53.54 & 74.57 \\
\rowcolor{tableours}[\tabcolsep][\tabcolsep]\textbf{BWM} & \textbf{67.76} & \underline{58.18} & \textbf{40.10} & \textbf{92.32} & \underline{16.90} & 4.09 & 49.74 & \underline{69.49} & \underline{70.53} & \underline{75.08} \\
\rowcolor{tablegroupwarm}[\tabcolsep][\tabcolsep]\multicolumn{11}{l}{\emph{Wrist view}} \\
Cosmos-Predict 2.5 & \underline{64.25} & \underline{50.15} & \underline{34.34} & \textbf{94.22} & \underline{66.60} & 90.51 & \textbf{82.46} & 79.79 & \underline{91.29} & \underline{4.14} \\
Ctrl-World & 51.59 & 44.44 & 26.53 & 36.55 & 39.64 & 79.11 & 64.24 & 75.46 & \textbf{91.48} & 2.47 \\
WorldGym & 62.12 & 44.91 & 32.64 & 74.17 & \textbf{69.35} & \textbf{93.33} & 79.38 & \underline{79.98} & 90.70 & 2.45 \\
\rowcolor{tableours}[\tabcolsep][\tabcolsep]\textbf{BWM} & \textbf{64.39} & \textbf{50.16} & \textbf{36.27} & \underline{93.77} & 66.45 & \underline{91.68} & \underline{80.79} & \textbf{80.19} & 90.88 & \textbf{4.79} \\
\bottomrule
\end{tabularx}
\end{minipage}
}
\end{table}

\begin{table}[!t]
\centering
\caption{Simulator performance on physical-robot videos, covering physics adherence, 3D accuracy, and controllability from third-person and wrist-camera views. ``--'' denotes unavailable results for action following and wrist-view trajectory accuracy.}
\label{tab:real_video_quality_physical_control}
{\renewcommand{\arraystretch}{1.10}
\begin{minipage}{0.98\linewidth}
\begin{tabularx}{\linewidth}{>{\raggedright\arraybackslash}p{0.25\linewidth}*{7}{Y}}
\toprule
\multicolumn{1}{c}{\multirow{2}{*}{Model}} & \multicolumn{2}{c}{\makebox[0pt][c]{Physics Adherence}} & \multicolumn{2}{c}{3D Accuracy} & \multicolumn{3}{c}{Controllability} \\
\cmidrule(lr){2-3}\cmidrule(lr){4-5}\cmidrule(lr){6-8}
& IntQ & TrajA & DepA & Persp & InstF & SemA & ActF \\
\midrule
\rowcolor{tablegroupwarm}[\tabcolsep][\tabcolsep]\multicolumn{8}{l}{\emph{Third-person view}} \\
Cosmos-Predict 2.5 & \underline{71.33} & \underline{94.10} & 98.99 & \underline{99.67} & \underline{79.33} & \underline{91.38} & -- \\
Ctrl-World & 48.67 & 26.80 & 89.03 & 92.00 & 25.67 & 78.46 & -- \\
WorldGym & 63.67 & 79.10 & \underline{99.33} & \textbf{100.00} & 64.33 & 90.55 & -- \\
\rowcolor{tableours}[\tabcolsep][\tabcolsep]\textbf{BWM} & \textbf{72.00} & \textbf{96.12} & \textbf{99.53} & \textbf{100.00} & \textbf{79.67} & \textbf{92.60} & -- \\
\rowcolor{tablegroupwarm}[\tabcolsep][\tabcolsep]\multicolumn{8}{l}{\emph{Wrist view}} \\
Cosmos-Predict 2.5 & \underline{35.67} & -- & \underline{95.19} & \underline{62.67} & \underline{24.67} & \underline{87.77} & -- \\
Ctrl-World & 23.33 & -- & 80.11 & 60.67 & 20.00 & 78.28 & -- \\
WorldGym & 32.33 & -- & 93.12 & \textbf{66.00} & \textbf{25.00} & 86.31 & -- \\
\rowcolor{tableours}[\tabcolsep][\tabcolsep]\textbf{BWM} & \textbf{36.67} & -- & \textbf{96.03} & 61.67 & 23.67 & \textbf{88.45} & -- \\
\bottomrule
\end{tabularx}
\end{minipage}
}
\end{table}

\endgroup

BWM obtains the \textbf{highest EWMScore in both views}, reaching \textbf{67.76} over 15 available metrics for the third-person view and \textbf{64.39} over 14 available metrics for the wrist view.
These scores exceed the strongest baseline (Cosmos-Predict~2.5) by 1.23 in the third-person view and 0.14 in the wrist view.
In the third-person view, BWM leads aesthetic quality, JEPA similarity, interaction quality, trajectory accuracy, depth accuracy, instruction following, and semantic alignment, while tying the best perspectivity.
In the wrist-camera view, it leads image quality, aesthetic quality, subject consistency, photometric consistency, interaction quality, depth accuracy, and semantic alignment.
The results demonstrate consistent advantages on key simulator-fidelity metrics across third-person and wrist-camera observations.

\subsubsection{Data Engine}

The physical data-engine study measures whether BWM-generated trajectories improve policy learning.
The downstream MOTIF policy~\citep{motif2026} is trained on 50 real trajectories per task.
Each augmented policy retains the same real data and adds 40 trajectories generated by one world model.
Table~\ref{tab:real_data_engine} reports hardware success over 25 trials per task.

\begin{table}[H]
\centering
\caption{Data-engine performance on physical robots, covering success rates across six tasks. The reference policy uses 50 real trajectories per task, while each augmented policy adds 40 trajectories from the named world simulator.}
\label{tab:real_data_engine}
{\renewcommand{\arraystretch}{1.10}
\setlength{\tabcolsep}{2.6pt}
\begin{tabular}{>{\raggedright\arraybackslash}p{0.256\linewidth}
  >{\centering\arraybackslash}p{0.070\linewidth}
  >{\centering\arraybackslash}p{0.070\linewidth}
  >{\centering\arraybackslash}p{0.070\linewidth}
  >{\centering\arraybackslash}p{0.105\linewidth}
  >{\centering\arraybackslash}p{0.130\linewidth}
  >{\centering\arraybackslash}p{0.125\linewidth}
  >{\centering\arraybackslash}p{0.065\linewidth}}
\toprule
\multicolumn{1}{c}{\multirow{2}{*}{Policy Training Data}} & Fold & Open & Stack & Push Ball & Push Block & Wipe & \multirow{2}{*}{Avg.} \\
& Towel & Drawer & Cups & to Target & off Platform & Whiteboard & \\
\midrule
\rowcolor{tablegroup}[\tabcolsep][\tabcolsep]Real only & 64 & 56 & 52 & 32 & 64 & 36 & 50.67 \\
Real + Ctrl-World & \underline{72} & 56 & \underline{60} & 36 & 56 & 40 & \underline{53.33} \\
Real + Cosmos-Predict 2.5 & 64 & \underline{64} & 56 & 32 & \underline{64} & 32 & 52.00 \\
Real + WorldGym & 28 & 40 & 52 & \underline{48} & \underline{64} & \underline{48} & 46.67 \\
\rowcolor{tableours}[\tabcolsep][\tabcolsep]\textbf{Real + BWM} & \textbf{88} & \textbf{68} & \textbf{72} & \textbf{56} & \textbf{84} & \textbf{58} & \textbf{71.00} \\
\bottomrule
\end{tabular}
}
\end{table}

\begin{figure}[H]
    \centering
    \includegraphics[width=0.62\linewidth]{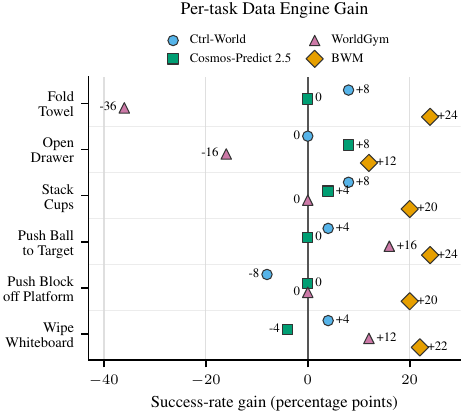}
    \caption{\textbf{Physical-robot Data Engine results.} Per-task gains are measured relative to real-only training and reproduce the results in Table~\ref{tab:real_data_engine}.}
    \label{fig:physical_data_engine}
\end{figure}

The BWM-augmented policy reaches the \textbf{highest mean hardware success rate of 71.00\%}.
It improves over real-only training by 20.33 and over Ctrl-World, the strongest evaluated world-simulator baseline, by 17.67.
BWM also gives the highest success rate on every task and is the only augmentation with positive gains across all six tasks.
Figure~\ref{fig:physical_data_engine} shows consistent improvements ranging from 12 to 24, demonstrating that BWM-generated trajectories provide effective policy supervision across the evaluated task set.

\subsubsection{Policy Evaluator}

\begin{figure}[!t]
    \centering
    \includegraphics[width=0.98\linewidth]{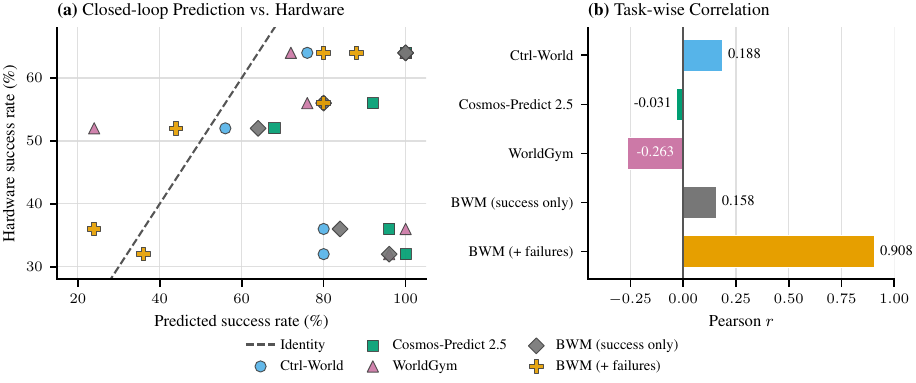}
    \caption{\textbf{Policy-evaluator agreement with hardware.} The left panel compares closed-loop predictions with hardware success rates across the six task pairs in Table~\ref{tab:real_policy_evaluator}. The right panel reports the corresponding Pearson correlations.}
    \label{fig:policy_evaluator_agreement}
\end{figure}

The final functional study tests whether BWM closed-loop rollouts predict the task-wise policy outcomes measured on hardware.
We alternate policy actions and BWM observation predictions for 25 rollouts per task.
The predicted success rate is the proportion of rollouts that satisfy the task-specific success criterion.
Mean Absolute Error (MAE) averages the six absolute task-wise errors, and $r$ is the Pearson correlation across the six paired task results.
Table~\ref{tab:real_policy_evaluator} compares these predictions with hardware outcomes.
Figure~\ref{fig:policy_evaluator_agreement} visualizes the task-wise agreement between closed-loop predictions and hardware outcomes.

\begin{table}[!t]
\centering
\caption{Task-wise policy-evaluator performance and agreement with hardware over 25 rollouts per task. Values in parentheses report the signed difference from the corresponding hardware success rate. Smaller absolute differences indicate closer task-wise agreement, with the smallest and second-smallest values shown in \textbf{bold} and \underline{underlined}, respectively. Lower MAE indicates better overall agreement, and the best and second-best MAE values follow the same notation.}
\label{tab:real_policy_evaluator}
{\renewcommand{\arraystretch}{1.10}
\setlength{\tabcolsep}{1pt}
\begin{tabular}{
  >{\raggedright\arraybackslash}m{0.202\linewidth}
  >{\centering\arraybackslash}m{0.085\linewidth}
  >{\centering\arraybackslash}m{0.085\linewidth}
  >{\centering\arraybackslash}m{0.085\linewidth}
  >{\centering\arraybackslash}m{0.100\linewidth}
  >{\centering\arraybackslash}m{0.120\linewidth}
  >{\centering\arraybackslash}m{0.108\linewidth}
  >{\centering\arraybackslash}m{0.070\linewidth}
  >{\centering\arraybackslash}m{0.070\linewidth}}
\toprule
\multicolumn{1}{c}{\multirow{2}{*}[-0.55em]{Evaluator}} &
\multicolumn{7}{c}{Task Success Rates (\%)} &
\multirow{2}{*}[-0.55em]{MAE$\downarrow$} \\
\cmidrule(lr){2-8}
&
\shortstack{Fold\\Towel} & \shortstack{Open\\Drawer} & \shortstack{Stack\\Cups} &
\shortstack{Push Ball\\to Target} & \shortstack{Push Block\\off Platform} &
\shortstack{Wipe\\Whiteboard} & Avg. & \\
\midrule
\rowcolor{tablegroup}[\tabcolsep][\tabcolsep]Hardware & 64 & 56 & 52 & 32 & 64 & 36 & 50.67 & -- \\
Ctrl-World & 76\,{\scriptsize(\underline{$+12$})} & 80\,{\scriptsize(\underline{$+24$})} & 56\,{\scriptsize($\mathbf{+4}$)} & 80\,{\scriptsize(\underline{$+48$})} & 100\,{\scriptsize(\underline{$+36$})} & 80\,{\scriptsize(\underline{$+44$})} & 78.67 & \underline{28.00} \\
Cosmos-Predict 2.5 & 100\,{\scriptsize($+36$)} & 92\,{\scriptsize($+36$)} & 68\,{\scriptsize($+16$)} & 100\,{\scriptsize($+68$)} & 100\,{\scriptsize(\underline{$+36$})} & 96\,{\scriptsize($+60$)} & 92.67 & 42.00 \\
WorldGym & 72\,{\scriptsize($\mathbf{+8}$)} & 76\,{\scriptsize($\mathbf{+20}$)} & 24\,{\scriptsize($-28$)} & 96\,{\scriptsize($+64$)} & 100\,{\scriptsize(\underline{$+36$})} & 100\,{\scriptsize($+64$)} & 78.00 & 36.67 \\
BWM (success only) & 100\,{\scriptsize($+36$)} & 80\,{\scriptsize(\underline{$+24$})} & 64\,{\scriptsize($+12$)} & 96\,{\scriptsize($+64$)} & 100\,{\scriptsize(\underline{$+36$})} & 84\,{\scriptsize($+48$)} & 87.33 & 36.67 \\
\rowcolor{tableours}[\tabcolsep][\tabcolsep]\textbf{BWM (+ failures)} & 88\,{\scriptsize($+24$)} & 80\,{\scriptsize(\underline{$+24$})} & 44\,{\scriptsize(\underline{$-8$})} & 36\,{\scriptsize($\mathbf{+4}$)} & 80\,{\scriptsize($\mathbf{+16}$)} & 24\,{\scriptsize($\mathbf{-12}$)} & 58.67 & \textbf{14.67} \\
\bottomrule
\end{tabular}
}
\end{table}

The BWM evaluator that includes failed rollouts reaches the {\bfseries\boldmath highest learned-evaluator correlation of $r = 0.908$}, together with the lowest MAE of 14.67.
Compared with Ctrl-World, BWM reduces MAE by 13.33 while increasing correlation by 0.720.
Including failed rollouts also raises correlation by 0.749 over the success-only variant and reduces its MAE by 22.00.
These results show that BWM is closest to hardware outcomes across both agreement measures.

\subsection{Ablation Studies}

These ablations examine how context construction, temporal design, action injection, conditioning signals, and observation resolution contribute to rollout fidelity, physical consistency, and action controllability.
Tables~\ref{tab:ablation_perceptual_metrics} and~\ref{tab:ablation_physical_control_metrics} report EWMScore and sixteen metrics spanning visual quality, motion quality, content consistency, physics adherence, 3D accuracy, and controllability.

\begin{table}[!t]
\centering
\caption{Ablation performance on the WorldArena benchmark, covering EWMScore, visual quality, motion quality, and content consistency. Metric abbreviations follow Table~\ref{tab:sim_video_quality_worldarena}. Light-blue rows indicate the selected configuration.}
\label{tab:ablation_perceptual_metrics}
{\renewcommand{\arraystretch}{1.02}
\setlength{\tabcolsep}{1pt}
\begin{minipage}{0.98\linewidth}
\begin{tabularx}{\linewidth}{>{\raggedright\arraybackslash}p{0.28\linewidth}*{10}{Y}}
\toprule
\multicolumn{1}{c}{\multirow{2}{*}{Setting}} & \multicolumn{1}{c}{\makebox[0pt][c]{Overall}} & \multicolumn{3}{c}{Visual Quality} & \multicolumn{3}{c}{Motion Quality} & \multicolumn{3}{c}{Content Consistency} \\
\cmidrule(lr){2-2}\cmidrule(lr){3-5}\cmidrule(lr){6-8}\cmidrule(lr){9-11}
& EWM & IQ & AQ & JS & DD & FS & MS & SC & BC & PC \\
\midrule
\rowcolor{tablegroup}[\tabcolsep][\tabcolsep]\multicolumn{11}{l}{\textit{Initial environment and dynamic history}} \\
None & 62.71 & 48.78 & 39.82 & 97.87 & 30.94 & 19.78 & 72.97 & 80.78 & 87.66 & \textbf{30.19} \\
Dynamic history & 62.77 & \textbf{48.87} & \textbf{40.08} & 97.74 & \textbf{31.64} & \textbf{20.00} & \textbf{73.93} & \textbf{81.71} & \textbf{88.35} & 30.07 \\
\rowcolor{tableours}[\tabcolsep][\tabcolsep]History + initial env. & \textbf{63.51} & 48.42 & 40.07 & \textbf{98.88} & 31.61 & 19.99 & 73.45 & 81.44 & 88.28 & 28.71 \\
\addlinespace[3pt]
\rowcolor{tablegroupwarm}[\tabcolsep][\tabcolsep]\multicolumn{11}{l}{\textit{Future prediction length}} \\
32 frames & 63.15 & 48.27 & 39.66 & 98.86 & 31.45 & 19.75 & 73.03 & 81.08 & 87.69 & \textbf{28.82} \\
\rowcolor{tableours}[\tabcolsep][\tabcolsep]72 frames & \textbf{63.51} & \textbf{48.42} & \textbf{40.07} & \textbf{98.88} & \textbf{31.61} & \textbf{19.99} & \textbf{73.45} & \textbf{81.44} & \textbf{88.28} & 28.71 \\
\addlinespace[3pt]
\rowcolor{tablegroup}[\tabcolsep][\tabcolsep]\multicolumn{11}{l}{\textit{History-window length}} \\
16 frames & 63.02 & \textbf{49.08} & 39.94 & 97.68 & 31.50 & 19.48 & 72.99 & 82.18 & 88.96 & \textbf{29.34} \\
4 frames & 55.49 & 46.89 & \textbf{40.21} & 54.09 & \textbf{31.61} & 15.90 & 71.79 & \textbf{84.48} & \textbf{89.42} & 25.86 \\
\rowcolor{tableours}[\tabcolsep][\tabcolsep]8 frames & \textbf{63.51} & 48.42 & 40.07 & \textbf{98.88} & \textbf{31.61} & \textbf{19.99} & \textbf{73.45} & 81.44 & 88.28 & 28.71 \\
\addlinespace[3pt]
\rowcolor{tablegroupwarm}[\tabcolsep][\tabcolsep]\multicolumn{11}{l}{\textit{Action-injection training configuration}} \\
Noise & 63.10 & 48.75 & 39.97 & 96.91 & 31.41 & 19.84 & 72.37 & 81.03 & 87.49 & 28.72 \\
AdaLN & 61.12 & 48.80 & 40.01 & 94.97 & 30.04 & 17.52 & 70.89 & \textbf{81.85} & 87.58 & \textbf{30.79} \\
Cross-attention & 59.54 & \textbf{49.07} & 40.04 & 91.06 & 28.95 & 17.26 & 71.79 & 80.17 & 86.29 & 29.26 \\
\rowcolor{tableours}[\tabcolsep][\tabcolsep]AdaLN + Cross-attention & \textbf{63.51} & 48.42 & \textbf{40.07} & \textbf{98.88} & \textbf{31.61} & \textbf{19.99} & \textbf{73.45} & 81.44 & \textbf{88.28} & 28.71 \\
\addlinespace[3pt]
\rowcolor{tablegroup}[\tabcolsep][\tabcolsep]\multicolumn{11}{l}{\textit{Conditioning signal}} \\
Text & 61.92 & 48.38 & 39.44 & 98.23 & 31.14 & 19.14 & 72.53 & 79.26 & 85.56 & 28.99 \\
Text+Action & 62.32 & \textbf{49.17} & \textbf{40.22} & 96.17 & 31.60 & 19.40 & 72.59 & \textbf{81.72} & \textbf{88.31} & \textbf{29.69} \\
\rowcolor{tableours}[\tabcolsep][\tabcolsep]Action & \textbf{63.51} & 48.42 & 40.07 & \textbf{98.88} & \textbf{31.61} & \textbf{19.99} & \textbf{73.45} & 81.44 & 88.28 & 28.71 \\
\addlinespace[3pt]
\rowcolor{tablegroupwarm}[\tabcolsep][\tabcolsep]\multicolumn{11}{l}{\textit{Observation-resolution processing}} \\
Native 240p & 50.36 & 25.98 & 32.40 & 69.80 & \textbf{32.87} & \textbf{35.16} & \textbf{73.78} & 76.42 & 86.89 & 26.10 \\
Resized to 480p & 56.38 & 34.22 & 35.90 & 91.12 & 28.01 & 16.46 & 68.85 & 77.50 & 83.70 & \textbf{36.02} \\
Super-resolved to 480p & 62.16 & \textbf{53.41} & \textbf{40.84} & 94.26 & 31.38 & 18.85 & 72.09 & \textbf{81.54} & 87.40 & 28.34 \\
\rowcolor{tableours}[\tabcolsep][\tabcolsep]Rerendered at 480p & \textbf{63.51} & 48.42 & 40.07 & \textbf{98.88} & 31.61 & 19.99 & 73.45 & 81.44 & \textbf{88.28} & 28.71 \\
\bottomrule
\end{tabularx}
\end{minipage}
}
\end{table}

\begin{table}[!t]
\centering
\caption{Ablation performance on the WorldArena benchmark, covering physics adherence, 3D accuracy, and controllability. Metric abbreviations follow Table~\ref{tab:sim_video_quality_physical_control}. Light-blue rows indicate the selected configuration.}
\label{tab:ablation_physical_control_metrics}
{\renewcommand{\arraystretch}{1.10}
\setlength{\tabcolsep}{2pt}
\begin{minipage}{0.98\linewidth}
\begin{tabularx}{\linewidth}{>{\raggedright\arraybackslash}p{0.38\linewidth}*{7}{Y}}
\toprule
\multicolumn{1}{c}{\multirow{2}{*}{Setting}} & \multicolumn{2}{c}{\makebox[0pt][c]{Physics Adherence}} & \multicolumn{2}{c}{3D Accuracy} & \multicolumn{3}{c}{Controllability} \\
\cmidrule(lr){2-3}\cmidrule(lr){4-5}\cmidrule(lr){6-8}
& IntQ & TrajA & DepA & Persp & InstF & SemA & ActF \\
\midrule
\rowcolor{tablegroup}[\tabcolsep][\tabcolsep]\multicolumn{8}{l}{\textit{Initial environment and dynamic history}} \\
None & 72.60 & 57.02 & 96.93 & 93.08 & 80.44 & \textbf{90.95} & \textbf{3.48} \\
Dynamic history & 71.84 & 56.65 & 96.19 & \textbf{93.44} & 79.72 & 90.93 & 3.13 \\
\rowcolor{tableours}[\tabcolsep][\tabcolsep]History + initial env. & \textbf{73.76} & \textbf{64.36} & \textbf{97.56} & 93.08 & \textbf{82.48} & 90.59 & 3.42 \\
\addlinespace[3pt]
\rowcolor{tablegroupwarm}[\tabcolsep][\tabcolsep]\multicolumn{8}{l}{\textit{Future prediction length}} \\
32 frames & \textbf{74.12} & 61.76 & 96.59 & \textbf{93.28} & 82.08 & \textbf{90.61} & 3.34 \\
\rowcolor{tableours}[\tabcolsep][\tabcolsep]72 frames & 73.76 & \textbf{64.36} & \textbf{97.56} & 93.08 & \textbf{82.48} & 90.59 & \textbf{3.42} \\
\addlinespace[3pt]
\rowcolor{tablegroup}[\tabcolsep][\tabcolsep]\multicolumn{8}{l}{\textit{History-window length}} \\
16 frames & 73.59 & 56.89 & 97.11 & \textbf{93.71} & 81.64 & \textbf{90.97} & 3.28 \\
4 frames & 58.36 & 41.43 & 94.14 & 84.44 & 57.12 & 90.55 & 1.53 \\
\rowcolor{tableours}[\tabcolsep][\tabcolsep]8 frames & \textbf{73.76} & \textbf{64.36} & \textbf{97.56} & 93.08 & \textbf{82.48} & 90.59 & \textbf{3.42} \\
\addlinespace[3pt]
\rowcolor{tablegroupwarm}[\tabcolsep][\tabcolsep]\multicolumn{8}{l}{\textit{Action-injection training configuration}} \\
Noise & \textbf{73.92} & 61.92 & 97.43 & \textbf{93.28} & \textbf{82.56} & \textbf{90.68} & 3.32 \\
AdaLN & 69.96 & 49.16 & 94.42 & 92.92 & 76.40 & 90.32 & 2.34 \\
Cross-attention & 68.10 & 43.42 & 89.89 & 90.94 & 73.79 & 90.36 & 2.23 \\
\rowcolor{tableours}[\tabcolsep][\tabcolsep]AdaLN + Cross-attention & 73.76 & \textbf{64.36} & \textbf{97.56} & 93.08 & 82.48 & 90.59 & \textbf{3.42} \\
\addlinespace[3pt]
\rowcolor{tablegroup}[\tabcolsep][\tabcolsep]\multicolumn{8}{l}{\textit{Conditioning signal}} \\
Text & 72.44 & 52.96 & 95.08 & \textbf{93.60} & 79.60 & \textbf{91.19} & 3.13 \\
Text+Action & 72.04 & 55.54 & 95.90 & 92.60 & 78.96 & 90.58 & 2.69 \\
\rowcolor{tableours}[\tabcolsep][\tabcolsep]Action & \textbf{73.76} & \textbf{64.36} & \textbf{97.56} & 93.08 & \textbf{82.48} & 90.59 & \textbf{3.42} \\
\addlinespace[3pt]
\rowcolor{tablegroupwarm}[\tabcolsep][\tabcolsep]\multicolumn{8}{l}{\textit{Observation-resolution processing}} \\
Native 240p & 47.32 & 18.82 & 79.74 & 68.36 & 40.64 & 88.20 & 3.20 \\
Resized to 480p & 59.44 & 39.97 & 93.32 & 81.44 & 62.32 & 90.50 & 3.38 \\
Super-resolved to 480p & 70.92 & 53.87 & 96.38 & \textbf{93.20} & 78.32 & \textbf{90.59} & 3.24 \\
\rowcolor{tableours}[\tabcolsep][\tabcolsep]Rerendered at 480p & \textbf{73.76} & \textbf{64.36} & \textbf{97.56} & 93.08 & \textbf{82.48} & \textbf{90.59} & \textbf{3.42} \\
\bottomrule
\end{tabularx}
\end{minipage}
}
\end{table}

\paragraph{Initial-environment guidance and dynamic history.}
Adding initial-environment guidance to dynamic history raises EWMScore from 62.77 to 63.51 and trajectory accuracy from 56.65 to 64.36.
This result indicates that initial-environment guidance preserves scene context while the dynamic history tracks recent changes during rollout.

\paragraph{Prediction chunk and history window.}
Increasing the prediction chunk from 32 to 72 frames raises EWMScore from 63.15 to 63.51 and trajectory accuracy from 61.76 to 64.36.
An eight-frame history gives the highest EWMScore and trajectory accuracy.
The sharp drop with four frames and the lower trajectory accuracy with sixteen frames indicate that the rollout benefits from a sufficiently informative but compact history.

\paragraph{Action-injection training configurations.}
Combining AdaLN with cross-attention gives the best EWMScore of 63.51 and trajectory accuracy of 64.36.
Compared with cross-attention alone, the combination improves these metrics by 3.97 and 20.94 while also outperforming AdaLN alone.
This result shows that jointly using AdaLN and cross-attention improves action conditioning.

\paragraph{Conditioning signal.}
Action conditioning reaches the best EWMScore of 63.51 and trajectory accuracy of 64.36, outperforming the Text and Text+Action settings.
The joint setting does not improve over action alone, suggesting that action is sufficient to specify the transition in this setting.

\paragraph{Observation-resolution processing.}
Rerendering at 480p improves EWMScore by 13.15 and trajectory accuracy by 45.54 over native 240p.
Although super-resolution attains the highest image quality, its trajectory accuracy remains 10.49 below rerendering, showing that visual enhancement alone does not replace action-aligned high-resolution replay.

\subsection{Qualitative Comparisons}
\begin{figure}[H]
    \centering
    \includegraphics[width=0.97\linewidth]{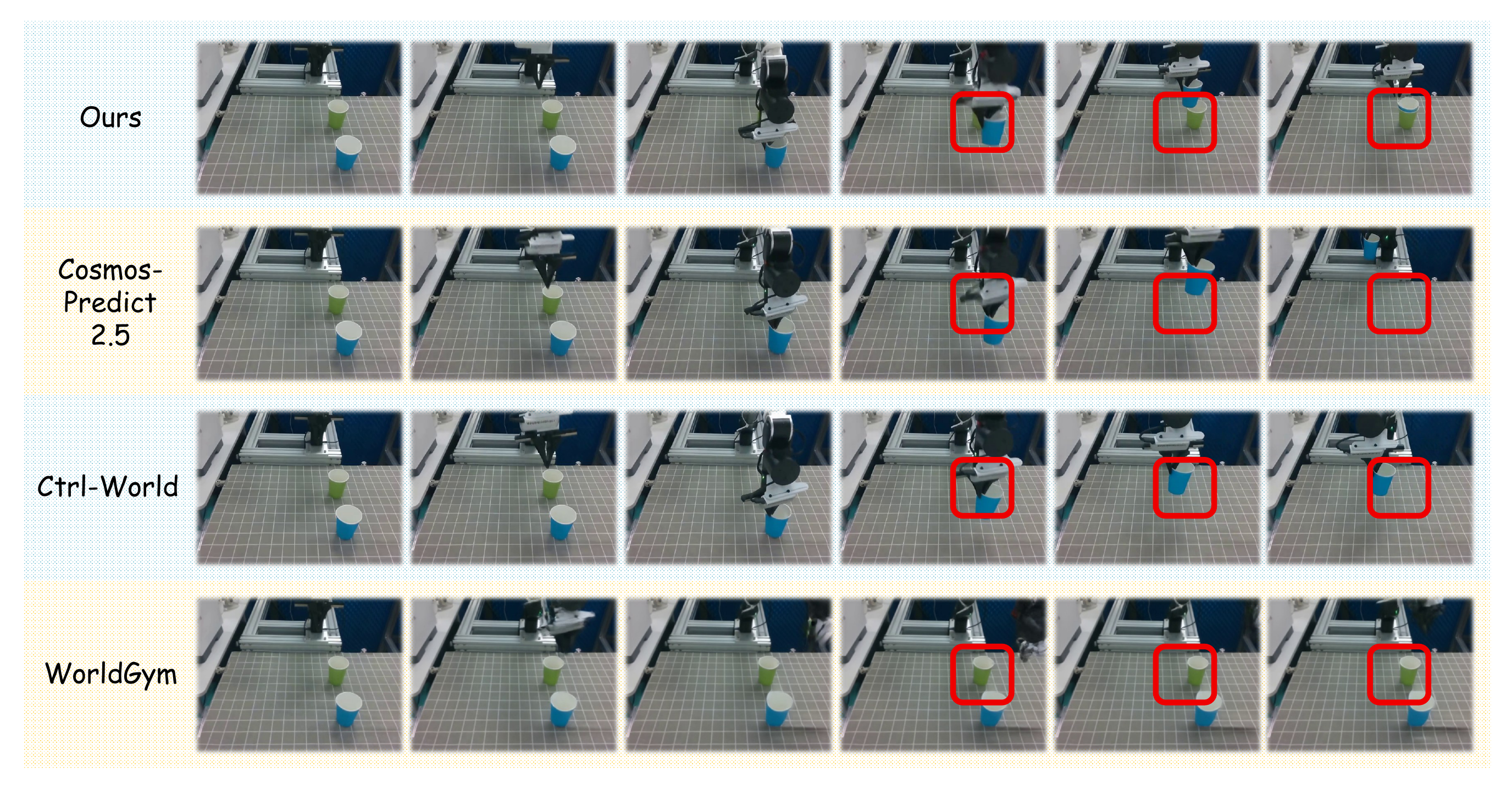}
    \caption{\textbf{Qualitative physical-robot data-engine case.} Each row progresses from left to right and shows a model-generated rollout. BWM is compared with Cosmos-Predict~2.5, Ctrl-World, and WorldGym. Red boxes highlight regions where the generated rollouts differ most clearly.}
    \label{fig:physical_robot_data_engine_case}
\end{figure}

\begin{figure}[H]
    \centering
    \includegraphics[width=0.97\linewidth]{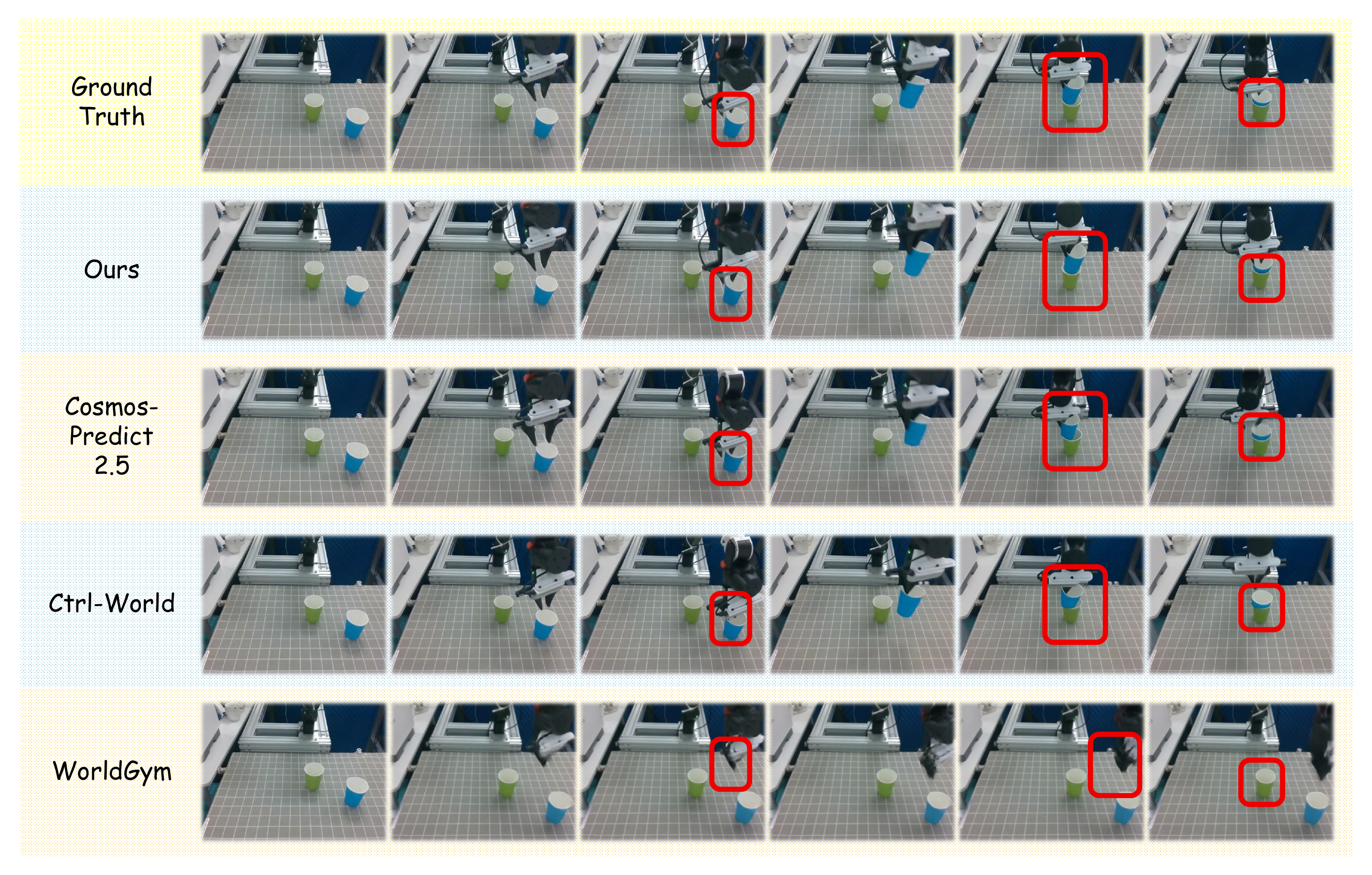}
    \caption{\textbf{Qualitative physical-robot policy-evaluator case.} Each row progresses from left to right. The hardware ground-truth sequence is followed by the rollouts predicted by BWM, Cosmos-Predict~2.5, Ctrl-World, and WorldGym. Red boxes highlight regions where the hardware and predicted sequences differ most clearly.}
    \label{fig:physical_robot_policy_evaluator_case}
\end{figure}

Figures~\ref{fig:physical_robot_data_engine_case} and~\ref{fig:physical_robot_policy_evaluator_case} present selected temporal predictions from the physical-robot stack cups task.
The data-engine case contrasts BWM with three world-model baselines, while the policy-evaluator case aligns the predicted rollouts with the hardware ground truth.
In these examples, BWM preserves both cups and advances the manipulated cup toward the demonstrated terminal state.
The baseline rollouts exhibit missing objects, incomplete interaction, or inconsistent terminal arrangements.

\section{Conclusion}

We present BWM, a low-cost, high-fidelity, action-conditioned world simulator that generates controllable robot rollouts from the current observation and a future action sequence. At the model level, BWM combines initial-environment guidance, dynamically updated context, and high-precision action control for stateful rollout generation. At the data level, trajectory replay, overlapping clip sampling, and initial-observation enhancement preserve action--observation alignment. We validate two representative functions of the simulator: a data engine for policy-training data and a policy evaluator for closed-loop assessment before hardware execution. Taken together, the simulation and physical-robot results show that BWM provides the strongest overall performance among the compared methods across simulator fidelity and the two validated functions, demonstrating broad capability in both high-fidelity simulation and functional utility. We release the model and code to support further development of functional robot world simulators.

\section*{Contributors}

\noindent\textbf{Main Contributors.}
Wentao Tan, Zengrong Lin, Bowen Wang, Zhe Li, Tianshi Wang, Yang Sun, Heng Zhi, Wenhao Liu, Zequn Wang, Enci Xie, Baixu Ji, Yipeng Chen, Chenyu Liu, Fengling Li, Lei Zhu$^{\dagger,\ddagger}$, Heng Tao Shen$^{\dagger}$.

\noindent\textbf{Other Contributors$^{\mathsection}$.}
Xuebin Fang, Chenming Li, Chen Xu, Hao Xue, Wenjie Yang, Pengfei Zhang, other members of the BWM Team.

\noindent{\footnotesize $^{\dagger}$ Corresponding authors. $^{\ddagger}$ Project leader. $^{\mathsection}$ Alphabetical order.}

\section*{Acknowledgments}

The computations in this work were supported by the Tongji University Intelligent Computing Center.

\bibliographystyle{unsrtnat}
\bibliography{references}

\clearpage
\appendix
\section{WorldArena Leaderboards}

This appendix presents the WorldArena leaderboards for BWM's base world-simulation capability and its two applications, the data engine and policy evaluator.
The method names displayed on the leaderboards do not fully match the BWM naming used in this paper.

\begin{figure}[H]
  \centering
  \includegraphics[width=\textwidth]{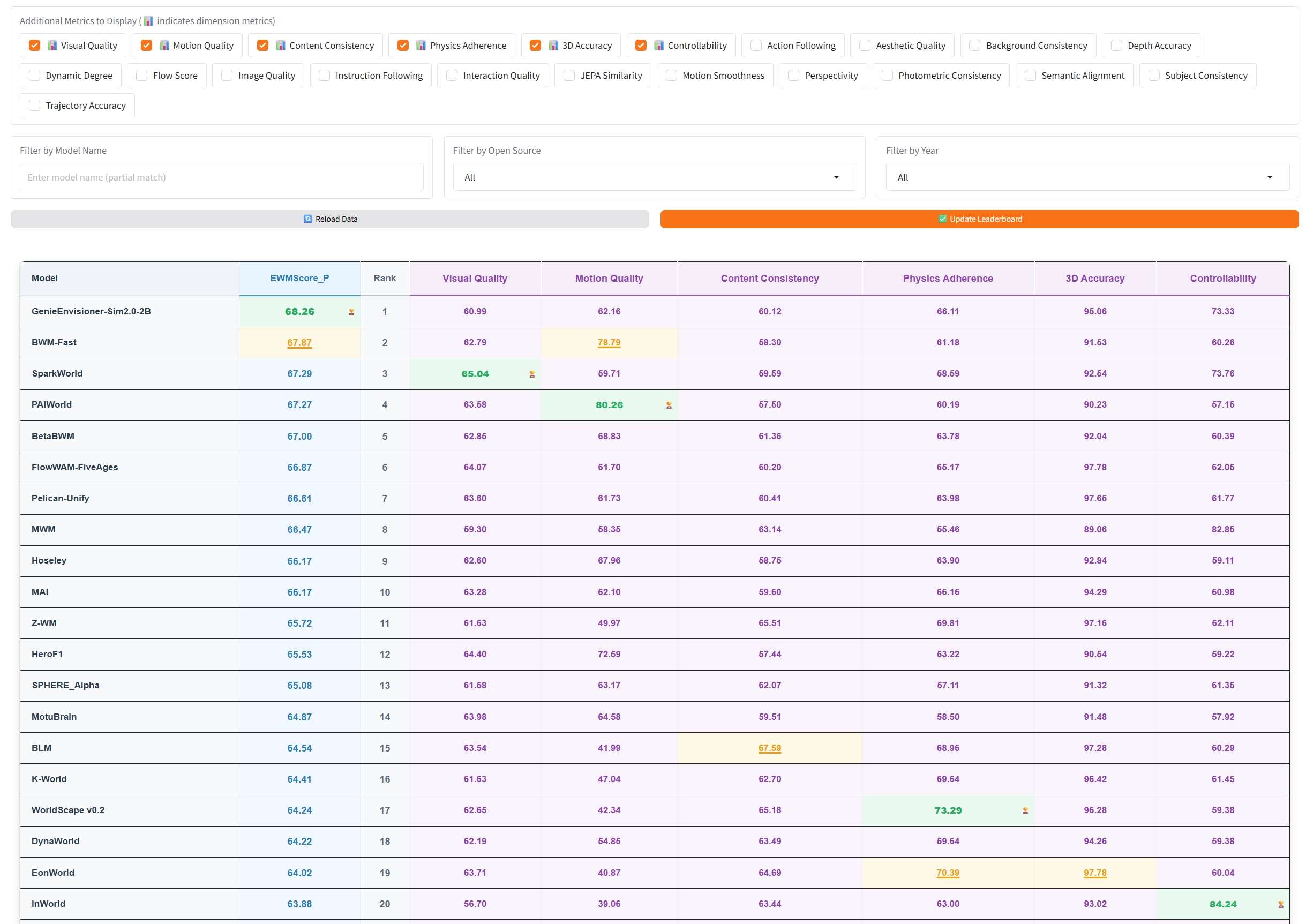}
  \caption{WorldArena Track 1 overall leaderboard. BWM-Fast ranks \textbf{second} by the overall metric.}
  \label{fig:appendix-track1-overall}
\end{figure}

\begin{figure}[H]
  \centering
  \includegraphics[width=\textwidth]{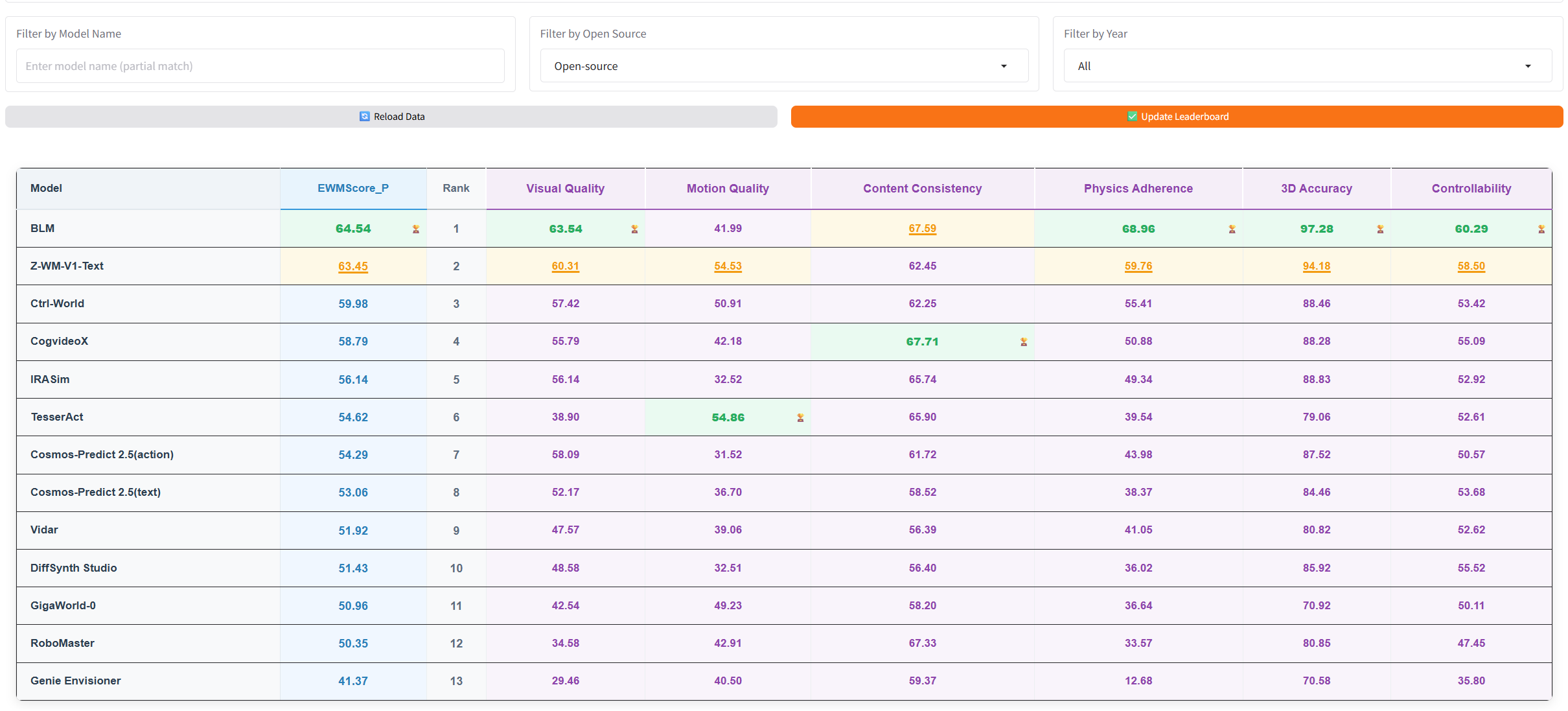}
  \caption{WorldArena Track 1 open-source leaderboard. BLM ranks \textbf{first} among the displayed open-source entries.}
  \label{fig:appendix-track1-open-source}
\end{figure}

\begin{figure}[H]
  \centering
  \includegraphics[width=\textwidth]{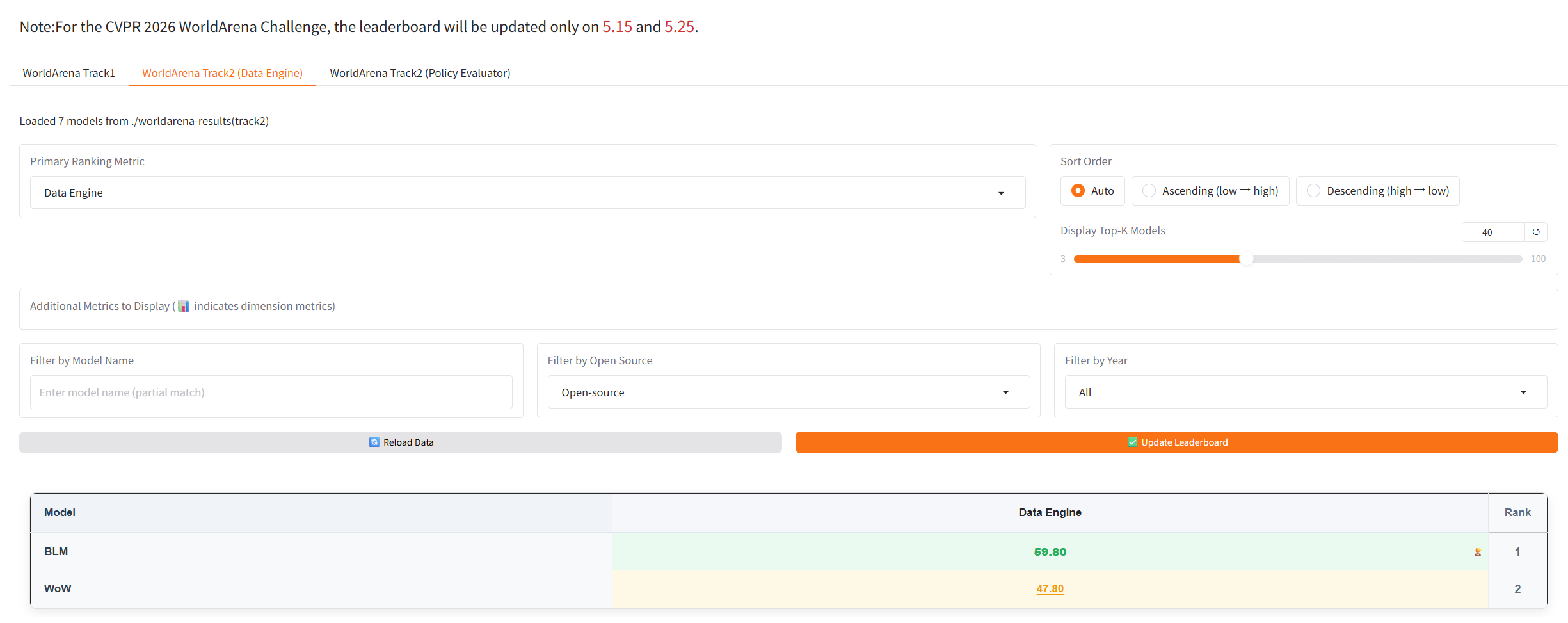}
  \caption{WorldArena Track 2 open-source Data Engine leaderboard. BLM ranks \textbf{first} among the displayed entries.}
  \label{fig:appendix-data-engine-open-source}
\end{figure}

\begin{figure}[H]
  \centering
  \includegraphics[width=\textwidth]{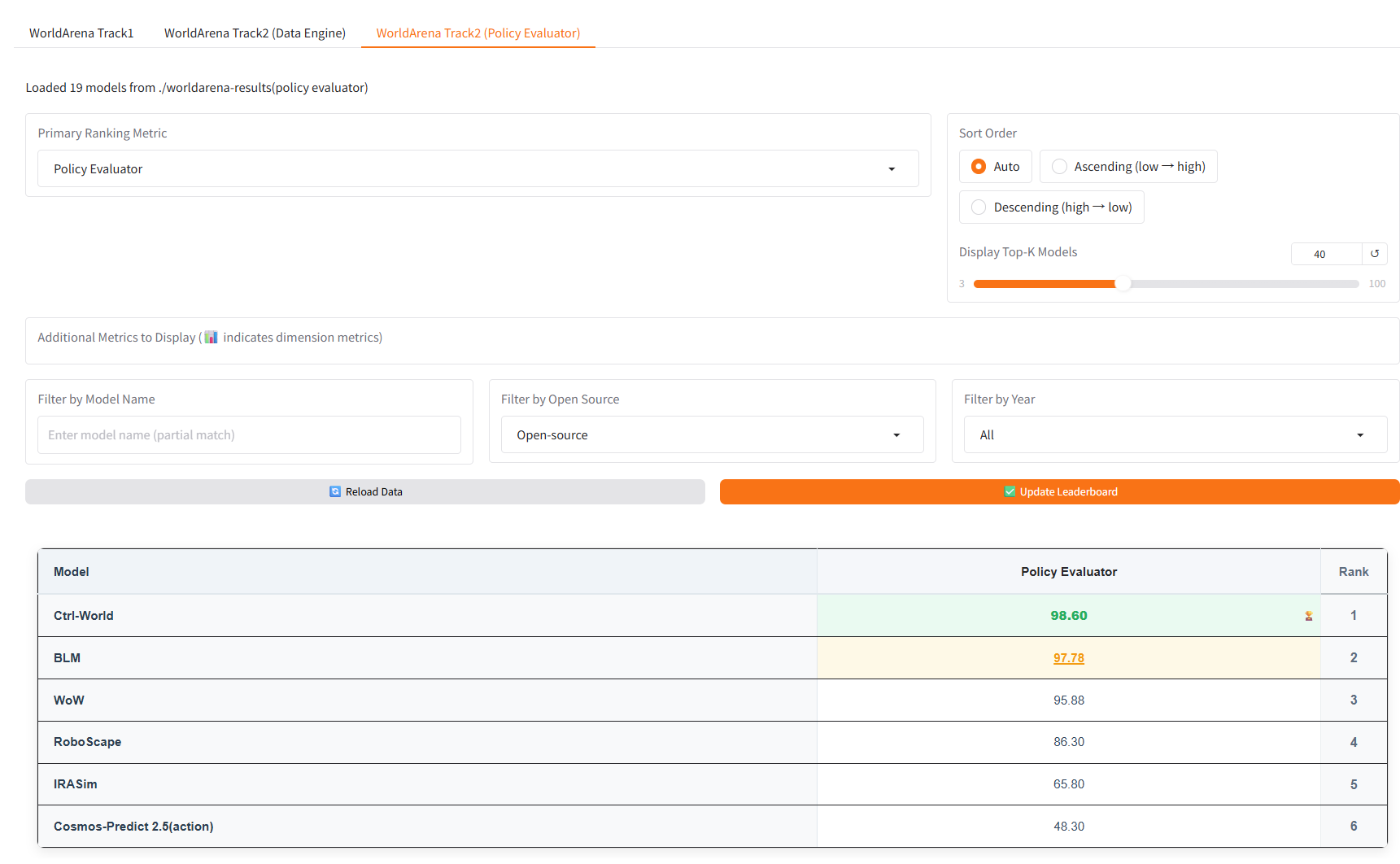}
  \caption{WorldArena Track 2 open-source Policy Evaluator leaderboard. BLM ranks \textbf{second} among the displayed entries.}
  \label{fig:appendix-policy-evaluator-open-source}
\end{figure}

\end{document}